\documentclass[11pt,a4paper]{article}   

\usepackage{graphicx}
\graphicspath{{}}

\usepackage[a4paper, left=0.99in, right=0.99in, top=1in, bottom=1in, includehead, includefoot, foot=0pt, head=0pt]{geometry} 
\parskip=\medskipamount 
\usepackage{tikz} 
\usetikzlibrary{arrows, arrows.meta} 
\usepackage{amssymb} 
\usepackage{amsmath} 
\usepackage{amsbsy} 
\usepackage{booktabs} 
\usepackage{multirow} 
\usepackage[scientific-notation=false, group-separator={,}]{siunitx} 
\usepackage{subfigure}
\usepackage[small,bf,hang]{caption}	
\usepackage{algorithm}
\usepackage[]{algcompatible} 
\usepackage[round, colon, authoryear]{natbib}
\usepackage{hyperref}
\hypersetup{
	colorlinks=true,
	citecolor=blue,
	linkcolor=teal,      
	urlcolor=blue,
}
\usepackage{titling} 
\usepackage{authblk} 

\usepackage{fancyhdr}
\usetikzlibrary{decorations.pathreplacing,calc}
\newcommand{\tikzmark}[1]{\tikz[overlay,remember picture] \node (#1) {};}
\newcommand*{\AddNote}[4]{%
	\begin{tikzpicture}[overlay, remember picture]
	\draw [decoration={brace,amplitude=0.5em, mirror}, decorate, thick,gray]
	($($(#3)-(0.1,0)$)!(#1.north)!($(#3)-(0.1,0)$)$) --  
	($($(#3)-(0.1,0)$)!(#2.west)!($(#3)-(0.1,0)$)$)
	node [align=center, text width=2.5cm, pos=0, anchor=east, rotate = 90, yshift = 12pt] {#4};
	\end{tikzpicture}
}%

\def\correspondingauthor{\footnote{Corresponding author: \href{mailto:roel.henckaerts@kuleuven.be}{roel.henckaerts@kuleuven.be}.}}

\begin{document}

\title{When stakes are high: balancing accuracy and transparency with Model-Agnostic Interpretable Data-driven suRRogates}
\author[,b,d]{Roel Henckaerts \correspondingauthor}
\author[b,c,d]{Katrien Antonio}
\author[a]{Marie-Pier C\^{o}t\'{e}}
\affil[a]{\'{E}cole d'actuariat, Universit\'{e} Laval, Canada.}
\affil[b]{Faculty of Economics and Business, KU Leuven, Belgium.}
\affil[c]{Faculty of Economics and Business, University of Amsterdam, The Netherlands.}
\affil[d]{LRisk, Leuven Research Center on Insurance and Financial Risk Analysis, KU Leuven, Belgium.}
\predate{}
\postdate{}
\date{}
\maketitle
\thispagestyle{empty}

\begin{abstract}
\noindent Highly regulated industries, like banking and insurance, ask for transparent decision-making algorithms.
At the same time, competitive markets are pushing for the use of complex black box models.
We therefore present a procedure to develop a Model-Agnostic Interpretable Data-driven suRRogate (maidrr) suited for structured tabular data.
Knowledge is extracted from a black box via partial dependence effects.
These are used to perform smart feature engineering by grouping variable values.
This results in a segmentation of the feature space with automatic variable selection.
A transparent generalized linear model (GLM) is fit to the features in categorical format and their relevant interactions.
We demonstrate our \textsf{R} package \texttt{maidrr} with a case study on general insurance claim frequency modeling for six publicly available datasets.
Our maidrr GLM closely approximates a gradient boosting machine (GBM) black box and outperforms both a linear and tree surrogate as benchmarks.
\\ [2mm]
\textbf{Key words}: Compliance, Feature selection, GLM, Insurance, Segmentation, XAI
\end{abstract}

\section{Introduction}
\label{intro}
The big data revolution opened the door to highly complex artificial intelligence (AI) technology in search for top performance. However, at the same time, there is growing public awareness for the issues of interpretability, explainability and fairness of AI systems \citep{Oneil2016}. The General Data Protection Regulation \citep{GDPR} introduces ``the right to an explanation'' of decision-making algorithms, thereby pushing for transparent communication on the underlying rationale of the decisions. An explainable AI (XAI) algorithm enables human users to understand, trust and manage its decisions \citep{Gunning2017}. Explainability is gaining attention in many industries, such as automotive \citep{Meteier2019}, banking \citep{Bracke2019}, healthcare \citep{Ahmad2018}, insurance \citep{OECD}, manufacturing \citep{Hrnjica2020} and critical systems \citep{Gade2019}. Full transparency is essential for high-stakes decisions with a big impact on a person's life, such as medical diagnosis, insurance coverage, education admission, loan applications, criminal justice, autonomous transportation and job recruitment.

A lack of algorithmic transparency can hinder AI implementations in business practice due to regulatory compliance requirements \citep{Arrieta2020}. XAI is therefore especially important in highly regulated industries with an extensive review of algorithms by supervisory authorities. Examples from the financial sector include the key information documents (KIDs) for packaged retail and insurance-based investment products \citep{PRIIPs}, detailed motivations for credit actions under the Equal Credit Opportunity Act \citep{ECOA} and filing requirements for general insurance rates to the National Association of Insurance Commissioners \citep{NAIC}. Our case study in Section~\ref{case} puts focus on general insurance pricing as one of the high-stakes XAI application areas where transparent decision-making is essential due to strict regulations.

A clear distinction regarding model explainability is made between interpretation techniques \emph{ex-post} and transparency \emph{ex-ante} \citep{Guidotti2018}. On the one hand, a wide range of interpretation techniques are available to aid users in the explainability of opaque models and their predictions \citep{Biecek2018}. On the other hand, decision trees, rules and linear models are transparent by design, meaning they are easily comprehensible for human users. In linear models, the contribution (sign and strength) of feature~$x_j$ to the prediction target~$y$ is directly observable from the model coefficient~$\beta_j$ \citep{Doran2017}. Furthermore, the output is simply visualized in a decision table, see Figure~\ref{lin_mod}. \citet{Huysmans2011} perform a user study on the comprehensibility of several representation formats and show that decision tables outperform trees and rules with respect to accuracy, response time, answer confidence and ease of use. 

\begin{figure}[h!]
	\begin{tabular}{cc}
		\begin{minipage}{0.64\textwidth}
			General formulation of a linear model: \\
			$\mathbb{E}[y] = \beta_0 + \beta_1 x_1 + \beta_2 x_2 + \cdots + \beta_p x_p$ \\ \\
			Return (\%) based on asset class and investment term: \\
			$\mathbb{E}[\texttt{return}] = 2 + 4 \, \texttt{asset}_{stock} + 3 \, \texttt{term}_{long}$
		\end{minipage}&
		\begin{minipage}{0.34\textwidth}
			\begin{tabular}{cc|c}
				\toprule
				\texttt{asset} & \texttt{term} & $\mathbb{E}$[\texttt{return}] \\
				\midrule
				bond & short & 2\% \\
				bond & long & 5\% \\
				stock & short & 6\% \\
				stock & long & 9\% \\
				\bottomrule
			\end{tabular}
		\end{minipage}
	\end{tabular}
	\caption{An example of a linear model (left) and the corresponding decision table (right).}
	\label{lin_mod}
\end{figure}

Surrogate models intend to copy the behavior of a complex system by capturing its essence in a simpler format. This is related to the ideas of model compression \citep{Bucila2006}, mimic learning \citep{Ba2014} and distillation \citep{Hinton2015}. These approaches transfer knowledge from a large/slow model into a compact/fast approximation, which can easily be deployed in environments with stringent space and time requirements. The underlying structure of the complex system is learned by using its predictions as labels for training the surrogate. Within XAI applications, an interpretable surrogate is used to explain the complex system. Global surrogates explain average model behavior for a given dataset \citep{Molnar2020b}. Local surrogates, such as LIME \citep{Ribeiro2016}, K-LIME \citep{Hall2017}, SHAP \citep{Lundberg2017}, Anchors \citep{Ribeiro2018} and SLIM \citep{Hu2020}, explain individual predictions by an interpretable model in the vicinity of the observation of interest.

This paper presents a procedure to develop a global surrogate for a complex system, with the goal of implementing the surrogate in production. The surrogate inherits the strengths of a sophisticated black box algorithm, delivered in a simpler format that is easier to understand, manage and implement. The resulting high degree of model transparency can boost AI business applications, especially in highly regulated sectors such as banking and insurance. Our procedure extracts knowledge from the complex system via ex-post interpretation techniques. Next, using these insights, it performs smart feature engineering on the training data. In the end, an ex-ante transparent surrogate is fit to the engineered training data. The surrogate closely approximates the black box model such that it can be used as a substitute with explanations readily available. 

We put forward the following three desirable properties. Firstly, a \emph{model-agnostic} procedure is preferred due to the ever increasing variety of black box algorithms. We rely on partial dependence (PD) effects to extract knowledge from the black box, thereby covering a vast amount of different model types \citep{Friedman2001}. Secondly, the resulting surrogate should be \emph{interpretable}, making it easy to comprehend and use by human users. We employ generalized linear models (GLMs), formulated by \citet{Nelder1972}. This versatile model class covers a broad range of classification and regression models and allows to represent its output as a decision table. GLMs are therefore widely used in for example the insurance industry. Thirdly, a \emph{data-driven} procedure avoids the need for ad hoc model choices. We fully automate the transformation from black box to transparent surrogate via a cross-validation scheme.

We introduce maidrr: a Model-Agnostic Interpretable Data-driven suRRogate procedure for a black box developed on structured tabular data. The complete procedure is available in the open source \textsf{R} package \texttt{maidrr} \citep{maidrr}. The rest of this paper is structured as follows. Section~\ref{metho} details the maidrr methodology. Section~\ref{case} shows an application to insurance claim frequency modeling, where transparency is essential due to strict regulations. We demonstrate that our maidrr surrogate GLM is able to approximate the performance of a black box closely, while outperforming a linear and tree benchmark surrogate. Section~\ref{concl} concludes this paper.

\section{Methodology}
\label{metho}
\begin{figure}[h]
	\begin{center}
		\resizebox{0.8\columnwidth}{!}{%
			\begin{tikzpicture}[->,>=stealth',auto,node distance=5cm,
			thick,main node/.style={font=\sffamily \bfseries \scriptsize}]
			
			\node[main node] (1)  at(0,0) {\includegraphics[width=0.075\textwidth]{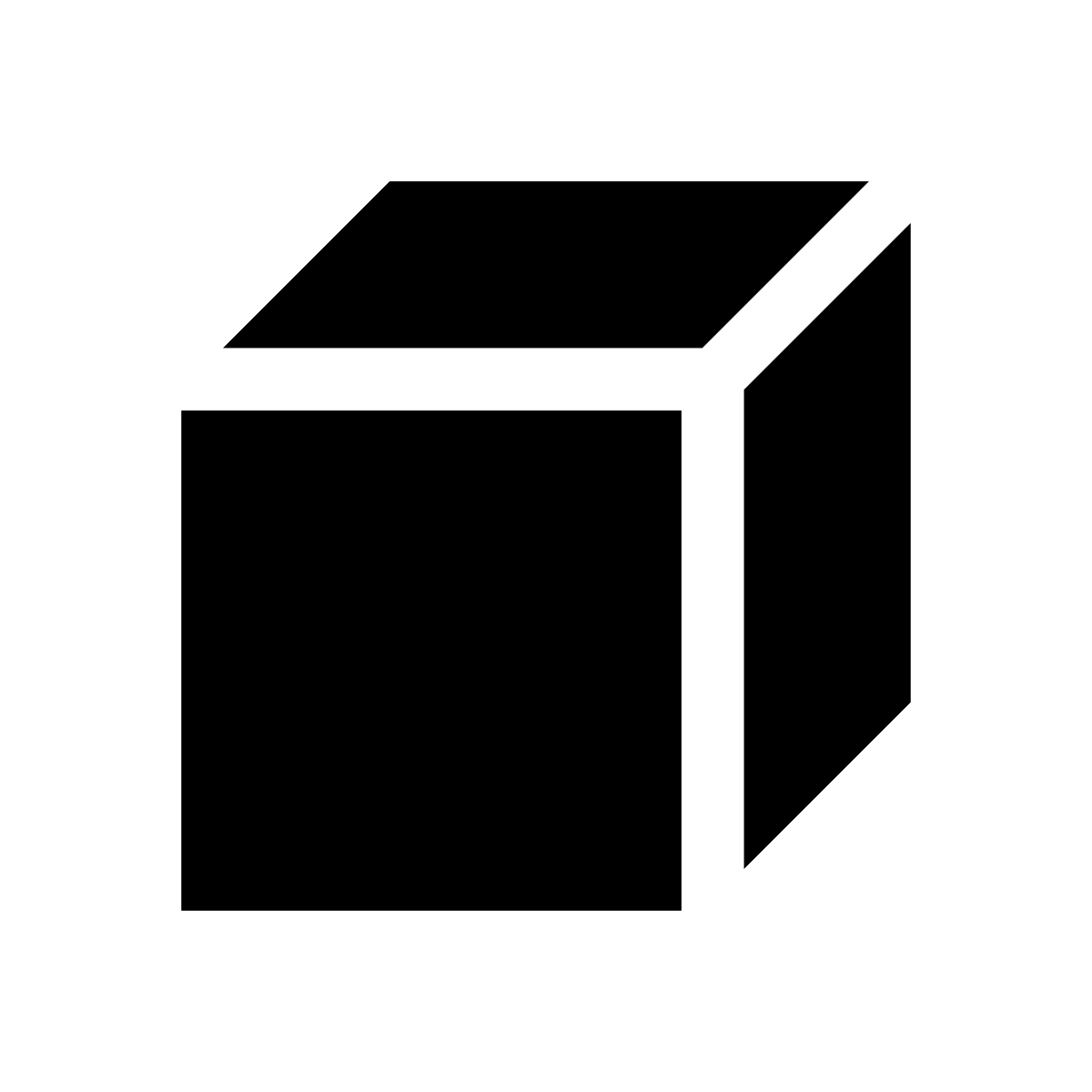}};
			\node[main node]  at(0,0.7) {Black box};
			
			\node[main node] (2)  at(3,0) {\includegraphics[width=0.08\textwidth]{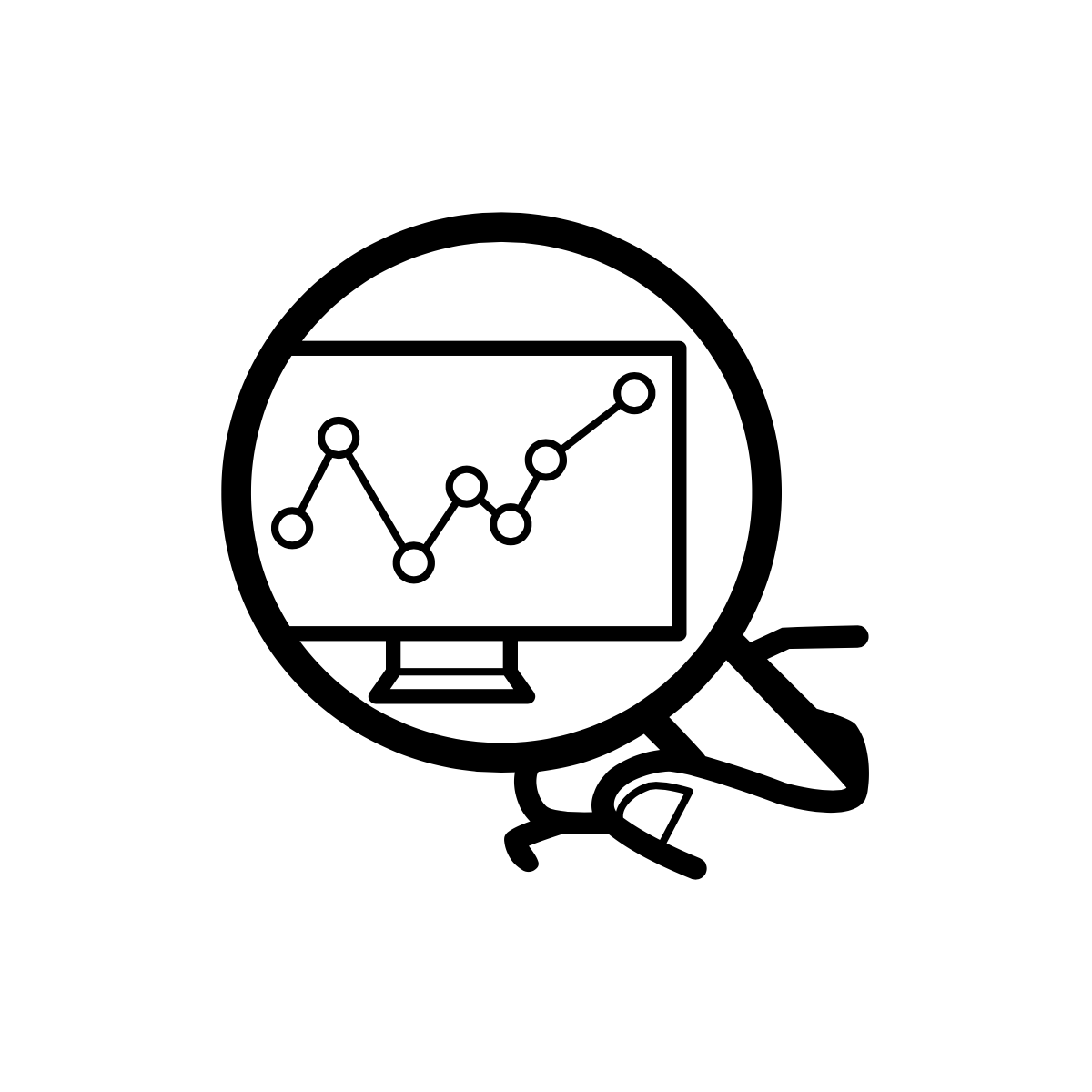}};
			\node[main node]  at(3,0.7) {Knowledge};
			
			\node[main node] (5)  at(6,0) {\includegraphics[width=0.08\textwidth]{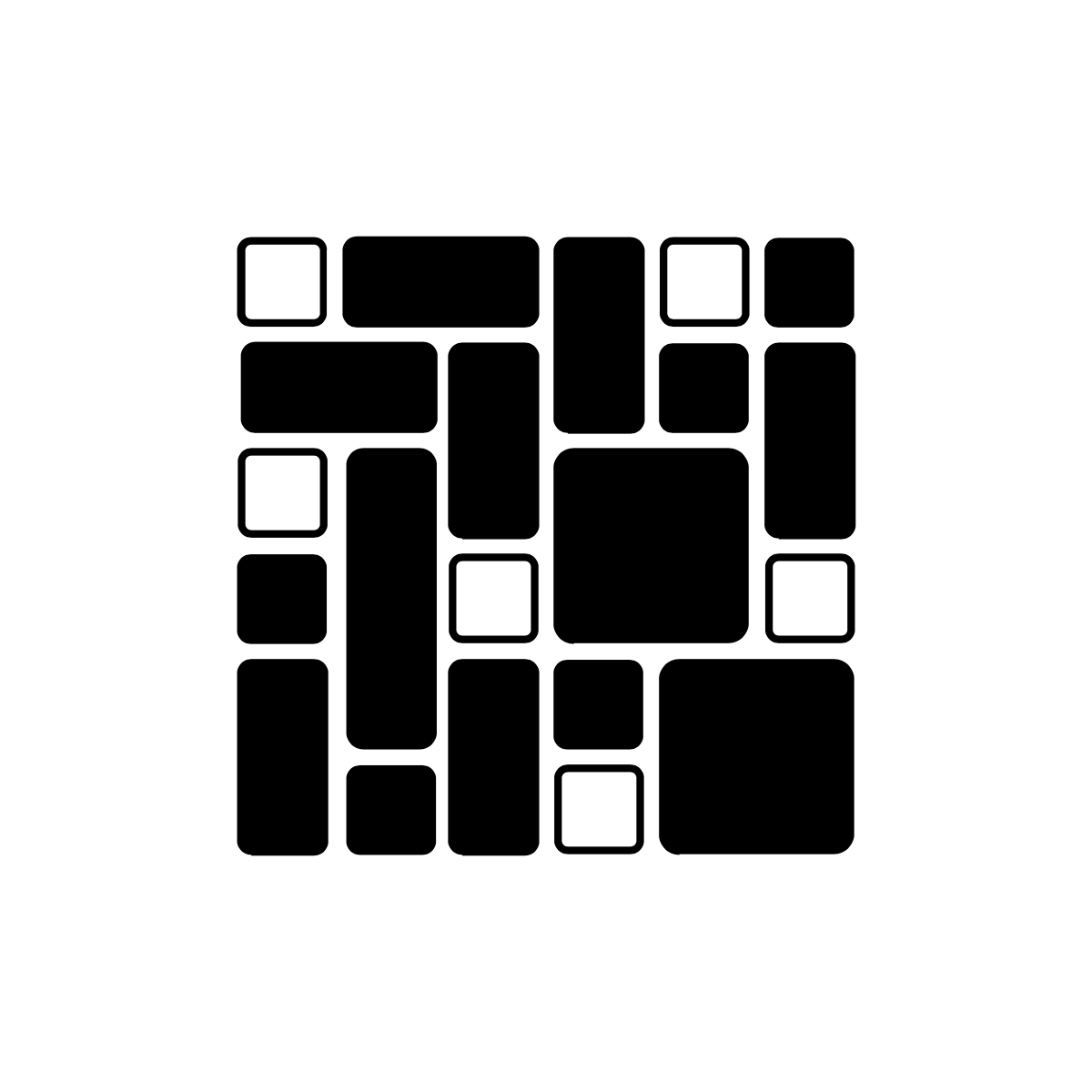}};
			\node[main node]  at(6,0.7) {Segmentation};
			
			\node[main node] (6)  at(9,0) {\includegraphics[width=0.08\textwidth]{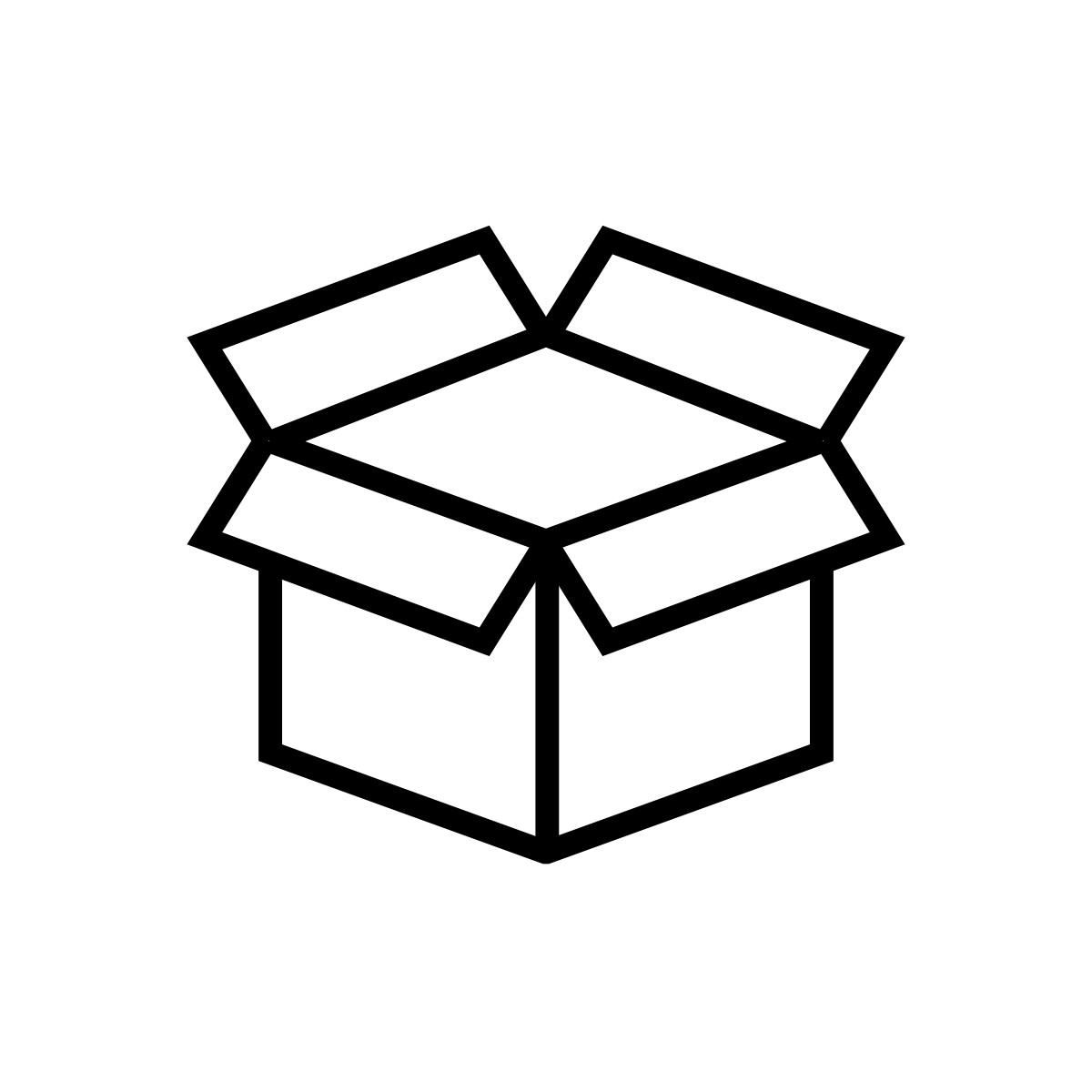}};
			\node[main node]  at(9,0.7) {Surrogate};
			
			\draw [->] (1.east) to node [midway,above] {PD} (2.west);
			\draw [->] (2.east) to node [midway,above] {DP} (5.west);
			\draw [->] (5.east) to node [midway,above] {GLM} (6.west);
			
			\end{tikzpicture}%
		}
		\caption{The maidrr process for transforming a black box algorithm into a transparent GLM.}
		\label{maidrr}
	\end{center}
\end{figure}
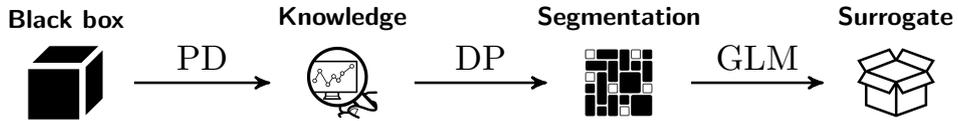

We first give an overview of the process behind maidrr, schematized in Figure~\ref{maidrr}. Afterward, we describe each step in details. The starting point is a black box that we want to transform into a simpler and more comprehensible surrogate. We extract knowledge from the black box in the form of partial dependence (PD) effects for all features involved. These PD effects, detailing the relation between a feature and the target, are used to group values/levels within a feature via dynamic programming (DP). A slightly different grouping approach is used for different types of features. For continuous/ordinal features, only adjacent values may be binned together, whereas any two levels within a nominal feature can be clustered. The binning/clustering via DP leads to an optimal and reproducible grouping of feature levels, resulting in a full segmentation of the feature space. After this step of feature engineering, a generalized linear model (GLM) is fit to the segmented data with all features in a categorical format and their relevant interactions. The end product is an interpretable surrogate which approximates the black box model.

\vspace{-1cm}

\paragraph{Black box} As a starting point, any black box model giving a prediction function~$f_{\textrm{pred}}(\boldsymbol{x})$ for features~$\boldsymbol{x} \in \mathbb{R}^p$ can be used. This property makes maidrr a model-agnostic procedure.

\paragraph{Knowledge} A univariate partial dependence (PD) captures the marginal relation between a feature~$x_j$, for $j\in\{1,\ldots,p\}$, and the model predictions \citep{Friedman2001}.  The PD effect~$\bar{f}_j(x_j)$ evaluates the prediction function~$f_{\textrm{pred}}$ for a given value of feature~$x_j$, while averaging over $n$ observed values of the other features~$\boldsymbol{x}^i_{-j}$ for observation~$i \in\{1,\ldots,n\}$:
\begin{equation}
\bar{f}_j(x_j) = \frac{1}{n} \sum_{i=1}^{n} f_{\textrm{pred}}(x_j,\boldsymbol{x}^i_{-j}).
\label{pdp_form_1D}
\end{equation}
The PD effect $\bar{f}_j$ is used to group values/levels within feature~$x_j$, as a similar PD indicates a similar relation to the prediction target. This grouping reduces the complexity of the feature with a limited loss of information. For feature~$x_j$, let $m_j$ denote the unique number of observed values and let $x_{j,q}$ denote its $q$th value for~$q\in\{1,\ldots,m_j\}$. We then define $z_{j,q}  = \bar{f}_j ( x_{j,q} )$ as the PD effect of feature~$x_j$ evaluated in~$x_{j,q}$. The goal is now to group the values~$x_{j,q}$ in $k_j$ groups based on~$z_{j,q}$. This represents a one-dimensional clustering problem of $z_{j,q}$ for $q \in \{1,\ldots,m_j\}$. In theory, PDs can be misleading for correlated features and accumulated local effects (ALE) serve as an alternative \citep{Apley2016}. However, Appendix~\ref{app_corr_feat} compares the resulting PDs and ALEs for highly correlated features, justifying the use of PDs for grouping purposes.

\paragraph{Segmentation} \citet{Wang2011} developed a dynamic programming (DP) algorithm for optimal and reproducible one-dimensional clustering problems. Elements of an $m_j$-dimensional input vector are assigned to $k_j$~clusters by minimizing the within-cluster sum of squares, that is, the sum of squared distances from each element to its corresponding cluster mean. This follows the same spirit as the classical $K$-means algorithm \citep{Macqueen1967}, but the DP algorithm guarantees reproducible and optimal groupings by progressively solving the sub-problem of clustering $u$~elements in $v$~clusters with $1 \leq u \leq m_j$ and $1 \leq v \leq k_j$. This algorithm is implemented in the \textsf{R} package \texttt{Ckmeans.1d.dp} \citep{Ckmeans.1d.dp} and allows for the inclusion of adjacency constraints in the clustering problem. We impose such constraints for continuous/ordinal features in order to group adjacent values. Nominal features are clustered without adjacency constraints such that any two levels can be grouped. The DP algorithm requires the specification of the number of groups~$k_j$ for feature~$x_j$. In theory, we can perform a $p$-dimensional grid search to find the optimal~$k_j$ for each feature~$x_j$ with $j \in\{1,\ldots,p\}$. However, this would cause the computation time to grow exponentially with~$p$, harming maidrr's scalability. We propose a penalized loss function to find the optimal number of groups~$k_j$.

\paragraph{Penalized loss function} After grouping feature~$x_j$ in $k_j$ groups, let $\tilde{z}_{j,q}$ represent the average PD effect for the group to which~$x_{j,q}$ belongs. We define a penalized loss function, which is to be minimized to find the optimal $k_j$ from a set of values, as follows:
\begin{equation}
\sum_{q=1}^{m_j} w_{j,q} \left(  z_{j,q} - \tilde{z}_{j,q} \right)^2  + \lambda \log(k_j).
\label{pen_loss}
\end{equation}
The first part of this loss function measures how well the PD effect is approximated by the grouped variant as a weighted mean squared error (wMSE) over all unique values of feature~$x_j$. The weight~$w_{j,q}$ represents the proportion of observations that equal value~$x_{j,q}$ for feature~$x_j$. This forces the procedure to focus on closely approximating frequently occurring feature values as opposed to rare cases. The second part of Eq.~\eqref{pen_loss} measures the complexity by means of the common logarithm of the number of groups~$k_j$. The penalty parameter~$\lambda$ acts as a bias-variance trade-off. A low (high) value of $\lambda$ allows for many (few) groups, resulting in an accurate (coarse) approximation of the PD. Note that $\lambda$ does not depend on $j$ in Eq.~\eqref{pen_loss}, which is adequate because the PD effects reside on the same scale, namely the scale of the predictions, see Eq.~\eqref{pdp_form_1D}. The original $p$-dimensional tuning problem in this way reduces to be one-dimensional over $\lambda$. The optimal~$\lambda$ value is determined via cross-validation, as detailed in Paragraph \nameref{hyperparam}.

\paragraph{Surrogate} Given a $\lambda$ value, we minimize Eq.~\eqref{pen_loss} for each of the features~$x_j$, resulting in a full segmentation of the feature space. After this step of feature engineering based on black box knowledge, we fit a transparent model to the original target and features in a categorical format. Generalized linear models (GLMs) allow for the specification of a diverse set of target distributions \citep{Nelder1972}. This facilitates the application of maidrr to classification tasks and many types of regression problems, for example linear, Poisson and gamma regression. We refer to Appendix~\ref{app_glm_form} for details on the GLM formulation. GLMs with only categorical features lead to fixed-size decision tables, see Appendix~\ref{app_glm_table} for an example. Even with many features they remain transparent, fileable in a tabular format and easy to use by business intermediaries, so the complexity of the GLM is not a concern. The high degree of transparency, thanks to observable coefficients, allows intuitive model post-processing by industry experts when necessary. GLMs are therefore attractive when transparency is essential, they are for example the preferred pricing tool within the strictly regulated insurance industry.

\paragraph{Feature interactions} So far we focused on grouping features via their marginal PDs, but interactions between features can play a major role in explaining the data. We first find a set of relevant interactions in the black box model by considering their strength as measured via the \mbox{$H$-statistic} \citep{Friedman2008}. Then, the pure interaction effect between features~$x_a$ and $x_b$ is captured by subtracting both one-dimensional PDs from the two-dimensional PD:
\begin{equation}
\bar{f}_{a,b}(x_a,x_b) = \frac{1}{n} \sum_{i=1}^{n} f_{\textrm{pred}}(x_a,x_b,\boldsymbol{x}^i_{-a,-b}) \, - \frac{1}{n} \sum_{i=1}^{n}  \sum_{\ell \in \{a,b\}}  f_{\textrm{pred}}(x_\ell,\boldsymbol{x}^i_{-\ell}).
\label{pdp_form_2D}
\end{equation}
We define feature~$x_{a:b}$ as the interaction containing all combinations of features~$x_a$ and~$x_b$ in the original data. The DP algorithm clusters levels in~$x_{a:b}$ that have similar~$\bar{f}_{a,b}(x_a,x_b)$ values, without any adjacency constraints. Interactions represent a correction on top of the marginal effects so we allow for maximum flexibility. Given a value of~$\lambda$, we determine the number of groups~$k_{ab}$ by minimizing the equivalent of Eq.~\eqref{pen_loss} obtained by computing the first term with Eq.~\eqref{pdp_form_2D}. The grouped version of~$x_{a:b}$ enters the surrogate GLM in a categorical format.

\paragraph{Hyperparameters} \label{hyperparam} Algorithm~\ref{maidrr_algo} details the full maidrr procedure with four input parameters: $\lambda_{\textrm{marg}}$, $\lambda_{\textrm{intr}}$, $k$ and $h$. A distinct value of~$\lambda$ is advised for marginal and interaction effects, as the PDs in Eq.~\eqref{pdp_form_1D} and \eqref{pdp_form_2D} reside on different scales. Marginal PDs are expressed on the scale of the predictions, whereas interaction PDs are expressed as a pure interaction effect. We tune the $\lambda$'s via $K$-fold cross-validation by iterating over a grid of~$\lambda$ values and choosing the optimal value that minimizes a loss function for the surrogate GLM predictions. This loss is computed with regards to the original data and not the black box predictions, resulting in a data-driven procedure. The tuning can be performed in two stages, first for~$\lambda_{\textrm{marg}}$ and next for~$\lambda_{\textrm{intr}}$, thereby avoiding a two-dimensional grid search. We refer to Section~\ref{surro_glm} for more details. Automatic feature selection is enabled as feature~$x_j$ is excluded from the surrogate when $k_j = 1$.  The hyperparameter~$k$ allows to specify a maximum number of groups for feature segmentation. The hyperparameter~$h$ selects a set of relevant interactions by means of a cut-off on the realized values of the black box's $H$-statistic, thereby excluding unimportant interactions upfront.

\vspace{2mm}

\begin{algorithm}[h!]
	\begin{algorithmic}
		\STATE\tikzmark{left}{\bfseries Input:} data, $f_{\textrm{pred}}$, $\lambda_{\textrm{marg}}$, $\lambda_{\textrm{intr}}$, $k$ and $h$
		\FOR{$j=1$ {\bfseries to} $p$} \tikzmark{top1} 
		\STATE calculate the PD effect $\bar{f}_j$ via Eq.~\eqref{pdp_form_1D}
		\STATE apply the DP algorithm to feature $x_j$ with $k_j^* = \underset{k_j \in \{1,\ldots,k\}}{\arg \min}$ Eq.~\eqref{pen_loss} for $\lambda =  \lambda_{\textrm{marg}}$
		\STATE $x_j^c$ represents the grouped version of $x_j$ in categorical format with $k_j^*$ groups
		\ENDFOR
		\STATE feature selection: $F = \{j \, \, | \, \, k_j^* > 1\}$ \tikzmark{bottom1}
		\STATE upfront interaction selection: $I = \{(l,m) \, \, | \, \, l \in F \, \, \textbf{and} \, \, m \in F \, \, \textbf{and} \, \, H(x_l,x_m) \geq h\}$ \tikzmark{top2} 
		\FORALL{$(a,b)$  {\bfseries in} $I$}
		\STATE calculate the PD effect $\bar{f}_{a,b}$ via Eq.~\eqref{pdp_form_2D}
		\STATE apply the DP algorithm to interaction $(x_a,x_b)$ with $k_{ab}^* = \underset{k_{j} \in \{1,\ldots,k\}}{\arg \min}$ Eq.~\eqref{pen_loss} for $\lambda =  \lambda_{\textrm{intr}}$
		\STATE $x_{a:b}^c$ represents the grouped version of $x_{a:b}$ in categorical format with $k_{ab}^*$ groups
		\ENDFOR
		\STATE interaction selection: $I = I \,\, \backslash \,\, \{(l,m) \, \, | \, \, k_{lm}^* = 1\}$ \tikzmark{bottom2} 
		\STATE fit a GLM to the target with features $x^c_j$ for $j \in F$ and interactions $x^c_{a:b}$ for $(a,b) \in I$
		\STATE {\bfseries Output:} surrogate GLM
		\AddNote{top1}{bottom1}{left}{marginal}
		\AddNote{top2}{bottom2}{left}{interaction}
	\end{algorithmic}
	\caption{maidrr}
	\label{maidrr_algo}
\end{algorithm}


\section{Case study for the insurance industry}
\label{case}
In most jurisdictions, insurers are required by law to document their pricing or rating model to the regulator. Determining a fair insurance quote is also high-stakes, as it can have a big impact on a person's life. This creates a clear need for transparency in the underlying decision-making process. A crucial part of ratemaking is the accurate modeling of the number of claims reported by a policyholder. We therefore apply maidrr to a general insurance claim frequency prediction problem. Section~\ref{data} introduces the model setting and the datasets. Section~\ref{surro} details the model construction for the black box and the maidrr GLM surrogate. Section~\ref{eval} evaluates the performance of the GLM with respect to the black box against two benchmark surrogates.

\subsection{Claim frequency modeling with insurance data}
\label{data}
We analyze six motor third party liability (MTPL) insurance portfolios, which are available in the \textsf{R} packages \texttt{CASdatasets} \citep{CASdatasets} or \texttt{maidrr} \citep{maidrr}. All datasets contain an MTPL portfolio followed over a period of one year, with the amount of policyholders~($n$) and the number of features~($p$) detailed in Table~\ref{datasets}. Each dataset holds a collection of different types of risk features, for example the age of the policyholder (continuous), the region of residence (nominal) and the type of insurance coverage (ordinal).

\begin{table}[h!]
	\centering
	\caption{Overview of the number of policyholders ($n$) and features ($p$) in the datasets\textsuperscript{*}.}
	\begin{tabular}{lcccccc}
		\toprule
		& \texttt{ausprivauto} & \texttt{bemtpl} & \texttt{freMPL} & \texttt{freMTPL} & \texttt{norauto} & \texttt{pricingame} \\
		\midrule
		$n$ & \num{67856} & \num{163210} & \num{137254} & \num{677925} & \num{183999} & \num{99859}  \\
		$p$ & 5 & 10 & 9 & 8 & 4 & 19 \\
		\bottomrule
		\multicolumn{7}{l}{\footnotesize \textsuperscript{*}The name of the dataset corresponds to its name in the \textsf{R} package.}
	\end{tabular}
	\label{datasets}
\end{table}

We model the number of claims filed during a given period of exposure-to-risk, defined as the fraction of the year for which the policyholder was covered by the insurance policy. Exposure is vital information, as filing one claim during a single month of coverage represents a higher risk than filing one claim during a full year. Table~\ref{nclaims} shows the distribution of the number of claims in the portfolios. Most policyholders do not file a claim, some file one claim and a small portion files two or more claims. Such count data is often modeled via Poisson regression, a specific form of GLM with a Poisson assumption for the target~$y$ and a logarithmic link function. In this setting, the industry standard is to incorporate the logarithm of exposure~$t$ via an offset term: $\ln(\mathbb{E}[y]) = \ln(t) + \beta_0 + \sum_j \beta_j x_j$. This leads to $\mathbb{E}[y] = t \times \exp(\beta_0 + \sum_j \beta_j x_j)$, that is, predictions are proportional to exposure and have a multiplicative structure: $\mathbb{E}[y] = t \times \exp(\beta_0) \times \prod_j  \exp(\beta_j x_j)$.

\begin{table}[h!]
	\centering
	\caption{Distribution of the number of claims in the portfolios.}
	\begin{tabular}{lrrrrrrr}
		\toprule
		& 0 & 1 & 2 & 3 & 4 & 5 & 6  \\
		\midrule
		\texttt{ausprivauto}  & \num{63232} & \num{4333} & \num{271} & \num{18} & \num{2} & \num{0} & \num{0} \\
		\texttt{bemtpl} & \num{144936} & \num{16539} & \num{1554} & \num{162} & \num{17} & \num{2} & \num{0} \\
		\texttt{freMPL}  & \num{106577} & \num{26068} & \num{4097} & \num{448} & \num{62} & \num{2} & \num{0} \\
		\texttt{freMTPL} & \num{643874} & \num{32175} & \num{1784} & \num{82} & \num{7} & \num{2} & \num{1}  \\
		\texttt{norauto} & \num{175555} & \num{8131} & \num{298} & \num{15} & \num{0} & \num{0} & \num{0}  \\
		\texttt{pricingame} & \num{87213} & \num{11232} & \num{1262} & \num{134} & \num{16} & \num{1} & \num{1}  \\
		\bottomrule
	\end{tabular}
	\label{nclaims}
\end{table}

\subsection{Finding a transparent model by opening the black box}
\label{surro}
Section~\ref{surro_bb} describes the construction of a gradient boosting machine or GBM as black box. Section~\ref{surro_glm}  details the maidrr procedure to obtain a GLM surrogate and illustrates the automatic feature selection and segmentation for several datasets.

\subsubsection{GBM as black box}
\label{surro_bb}
We opt for a gradient boosting machine or GBM \citep{Friedman2001} as the black box to start from. More specifically, we make use of stochastic gradient boosting \citep{Friedman2002} as implemented in the \textsf{R} package \texttt{gbm} \citep{gbm}. This choice is based on the good performance of GBMs as discussed in related work \citep{Henckaerts2020}. Due to the model-agnostic set up of maidrr, any model can be used as input, including deep neural networks.

We tune the number of trees~$T$ in the GBM via 5-fold cross-validation, see Table~\ref{gbm_ntree}. Other hyperparameters are fixed to a sensible value. Following \citet[Section~10.11]{Hastie2009}, we use decision trees of depth two, which are able to model up to third-order interactions. Each tree is built on randomly sampled data of size~$0.75 n$ and the learning rate is set to~$0.01$. To take into account the distributional characteristics of the count data, we use the Poisson deviance as loss function in the GBM tuning process. The Poisson deviance is defined as follows:

\begin{equation}
D^{\textrm{Poi}} \left\lbrace y, f_{\textrm{pred}}(\boldsymbol{x}) \right\rbrace = \frac{2}{n} \sum_{i=1}^{n} \left[  y_i \times \ln \left\lbrace  \frac{y_i}{f_{\textrm{pred}}(\boldsymbol{x}_i)} \right\rbrace  - \{y_i - f_{\textrm{pred}}(\boldsymbol{x}_i)\} \right] .
\label{poiss_dev}
\end{equation}

\begin{table}[h!]
	\centering
	\caption{Overview of the optimal number of trees ($T$) in the GBM for the different datasets.}
	\begin{tabular}{lcccccc}
		\toprule
		& \texttt{ausprivauto} & \texttt{bemtpl} & \texttt{freMPL} & \texttt{freMTPL} & \texttt{norauto} & \texttt{pricingame} \\
		\midrule
		$T$ & \num{474} & \num{3214} & \num{1377} & \num{3216} & \num{793} & \num{1198}  \\
		\bottomrule
	\end{tabular}
	\label{gbm_ntree}
\end{table}

\subsubsection{GLM surrogate via maidrr}
\label{surro_glm}
We build a surrogate GLM to approximate the optimal GBM for each dataset. The function \texttt{maidrr::autotune} \citep{maidrr} implements a tuning procedure for Algorithm~\ref{maidrr_algo}. 

Algorithm~\ref{maidrr_algo} requires four input parameters: $\lambda_{\textrm{marg}}$, $\lambda_{\textrm{intr}}$, $k$ and $h$. The $\lambda$ values determine the granularity of the resulting segmentation and GLM. We define a search grid for both~$\lambda$'s, ranging from $10^{-10}$ to $1$. This range is sufficiently wide for our application, as indicated by the optimal values in Table~\ref{lambdas}. Tuning of the $\lambda$ values is done in two stages. First, a grid search over $\lambda_{\textrm{marg}}$ finds the optimal GLM with only marginal effects by running the ``marginal'' part of Algorithm~\ref{maidrr_algo}. Then, a grid search over $\lambda_{\textrm{intr}}$ determines which interactions to include in that optimal GLM by running the ``interaction'' part of Algorithm~\ref{maidrr_algo}. This requires two one-dimensional grid searches of length $grid\_size$ instead of one two-dimensional search of length $grid\_size^2$, thereby saving computation time. The optimal $\lambda$ values are determined by performing 5-fold cross-validation on the resulting GLM with the Poisson deviance in Eq.~\eqref{poiss_dev} as loss function. The value of~$h$ determines the set of interactions that are considered for inclusion in the GLM by excluding meaningless interactions with a low $H$-statistic. This value is calculated automatically to consider the minimal set of interactions for which the empirical distribution function of the $H$-statistic exceeds 50\%. The intent is to take into account the most important interactions while still keeping the GLM simple. We set the maximum number of groups~$k=15$.

\begin{table}[h!]
	\centering
	\caption{Overview of the optimal $\lambda_{\textrm{marg}}$ and $\lambda_{\textrm{intr}}$ values for the different datasets.}
	\begin{tabular}{lcccccc}
		\toprule
		& \texttt{ausprivauto} & \texttt{bemtpl} & \texttt{freMPL} & \texttt{freMTPL} & \texttt{norauto} & \texttt{pricingame} \\
		\midrule
		$\lambda_{\textrm{marg}}$ & $4.2 \times 10^{-5}$ & $4.2 \times 10^{-5}$ & $1.6 \times 10^{-4}$ & $1.3 \times 10^{-7}$ & $1.1 \times 10^{-5}$  & $2.0 \times 10^{-6}$  \\
		$\lambda_{\textrm{intr}}$ & $8.5 \times 10^{-6}$ & $4.1 \times 10^{-6}$ & $4.6 \times 10^{-5}$ & $3.1 \times 10^{-6}$ & $3.1 \times 10^{-5}$  & $2.8 \times 10^{-6}$ \\
		\bottomrule
	\end{tabular}
	\label{lambdas}
\end{table}

\newpage 

\begin{figure}[h!]
	\centering
	\subfigure[Feature importance in the optimal GBM]{
		\includegraphics[width=0.42\textwidth]{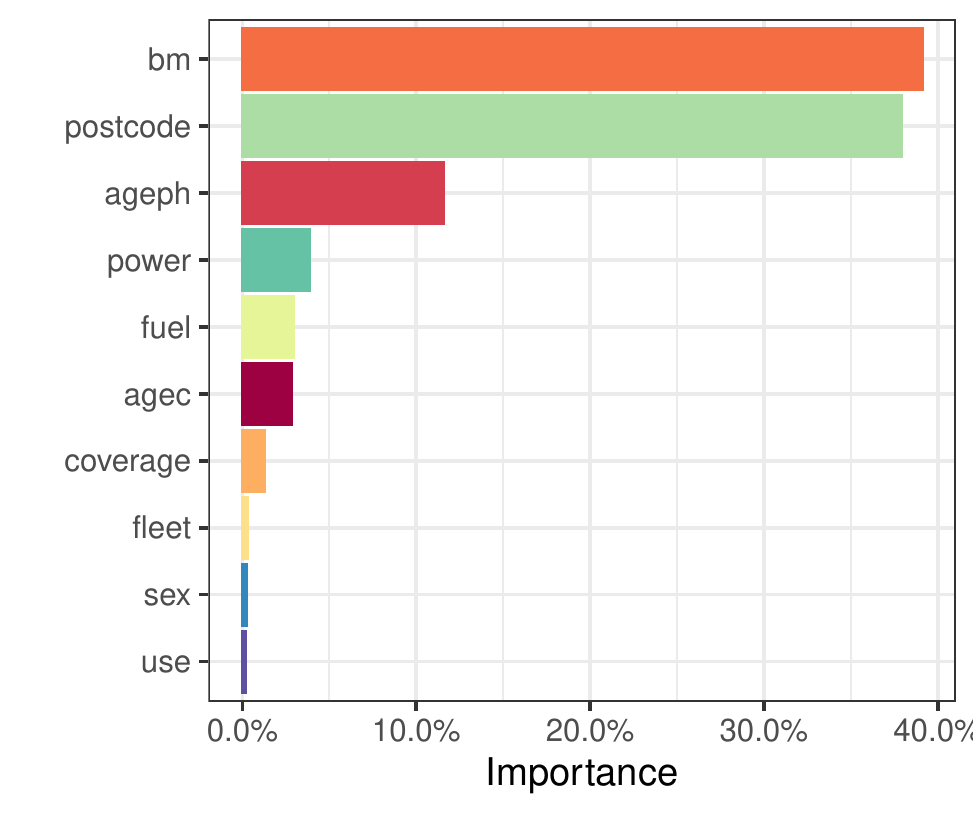}
		\label{feat_gbm}
	}
	\subfigure[Number of groups in function of $\lambda_{\textrm{marg}}$]{
		\includegraphics[width=0.525\textwidth]{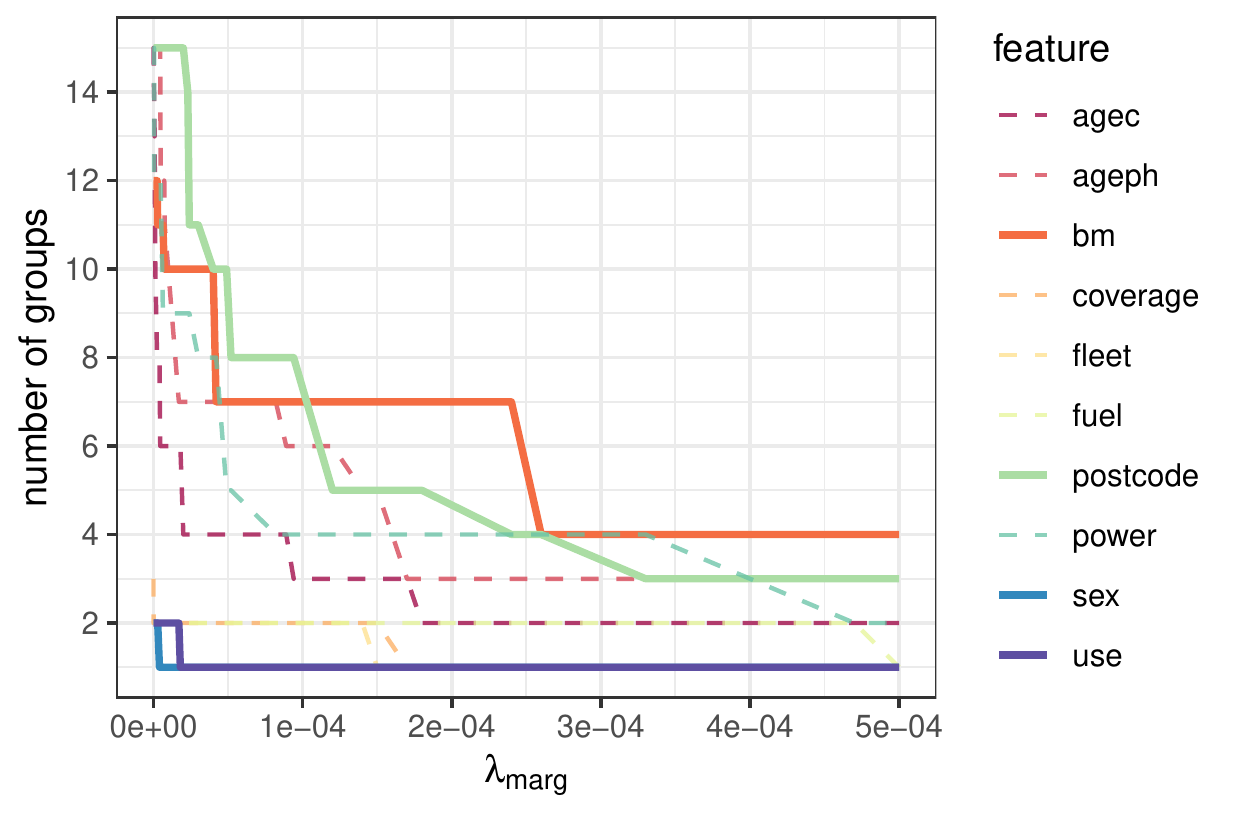}
		\label{lambda_slct}
	}
	\caption{Illustration of the automatic feature selection process in maidrr for \texttt{bemtpl}.}
	\label{auto_feat_slct}
\end{figure}

Figure~\ref{auto_feat_slct} illustrates the automatic feature selection of maidrr for the \texttt{bemtpl} portfolio. Figure~\ref{feat_gbm} shows feature importance scores according to the GBM and Figure~\ref{lambda_slct} shows the number of groups for each feature in function of~$\lambda_{\textrm{marg}}$. Important features, such as \texttt{bm} and \texttt{postcode}, retain a higher number of groups for increasing values of~$\lambda_{\textrm{marg}}$. Levels of uninformative features, like \texttt{use} and \texttt{sex}, are quickly placed in one group, effectively excluding these variables from the GLM. This is how maidrr performs automatic feature selection via the data-driven tuning of~$\lambda_{\textrm{marg}}$.  

\begin{figure}[ht!]
	\centering
	\subfigure[\texttt{bemtpl}: power of the vehicle in hp]{
		\includegraphics[width=0.475\textwidth]{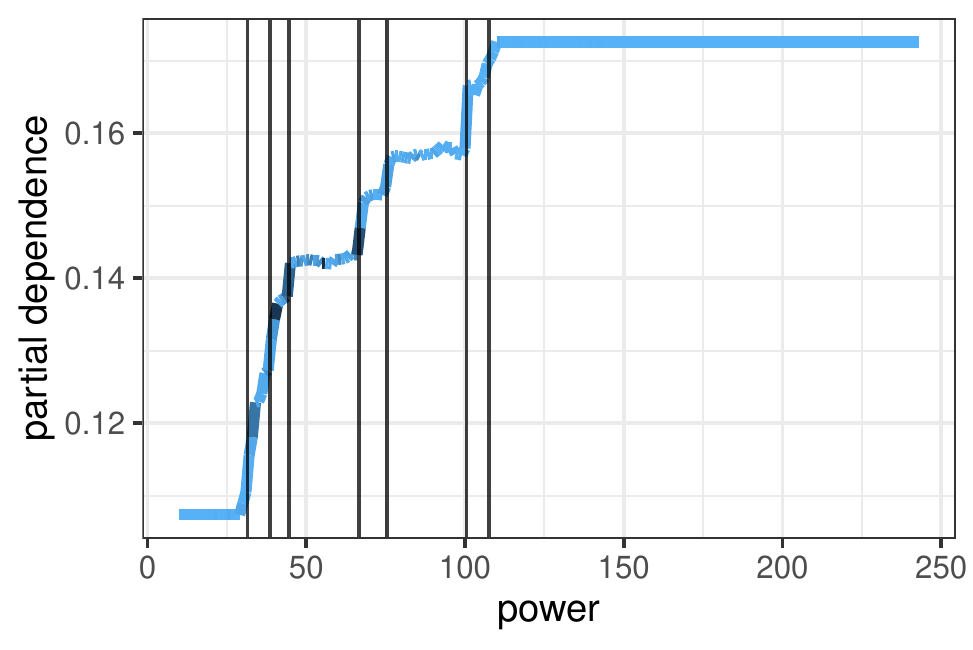}
		\label{bemtpl_power}
	}
	\subfigure[\texttt{pricingame}: age of the vehicle in years]{
		\includegraphics[width=0.475\textwidth]{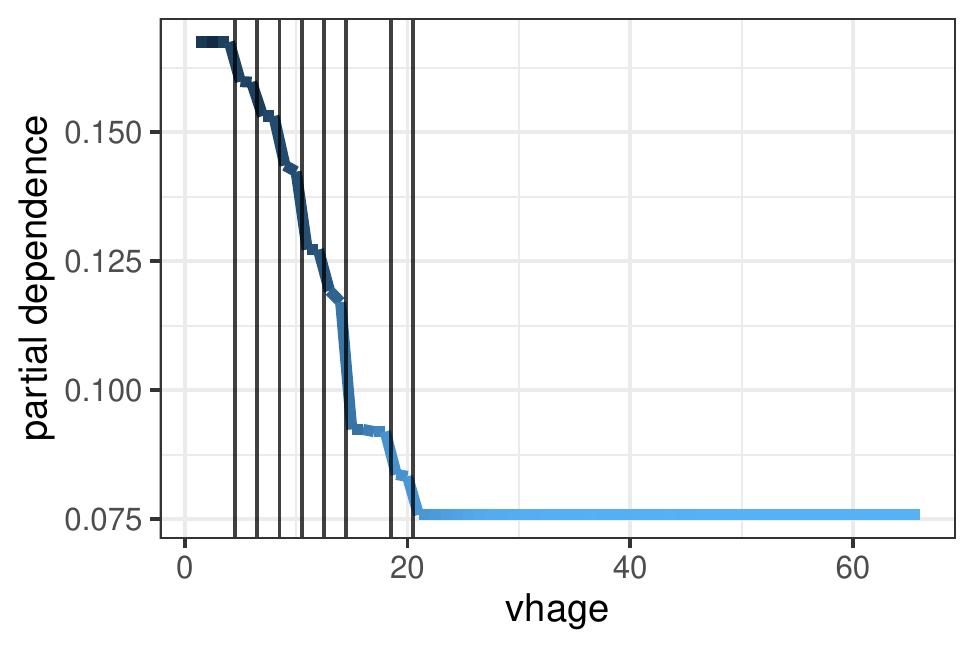}
		\label{pricingame_vhage}
	}
	\caption{PD effect and the resulting segmentation for two continuous features. Groups are separated by vertical lines and darker blue indicates a higher observation count in the portfolio.}
	\label{segm_cont}
\end{figure}

Figure~\ref{segm_cont} displays the resulting segmentation for two continuous features: vehicle power for \texttt{bemtpl} in Figure~\ref{bemtpl_power} and vehicle age for \texttt{pricingame} in Figure~\ref{pricingame_vhage}. Both show the GBM PD effect, where darker blue indicates a higher observation count in the portfolio. The features are grouped into \num{8} and \num{9} bins respectively, indicated by the vertical lines. The bins are wide wherever the PD effect is quite stable and narrow where the effect is steeper. We observe that claim risk increases for increasing vehicle power, while it decreases for increasing vehicle age.

\begin{figure}[hb!]
	\centering
	\subfigure[\texttt{ausprivauto}: age of the policyholder]{
		\includegraphics[width=0.475\textwidth]{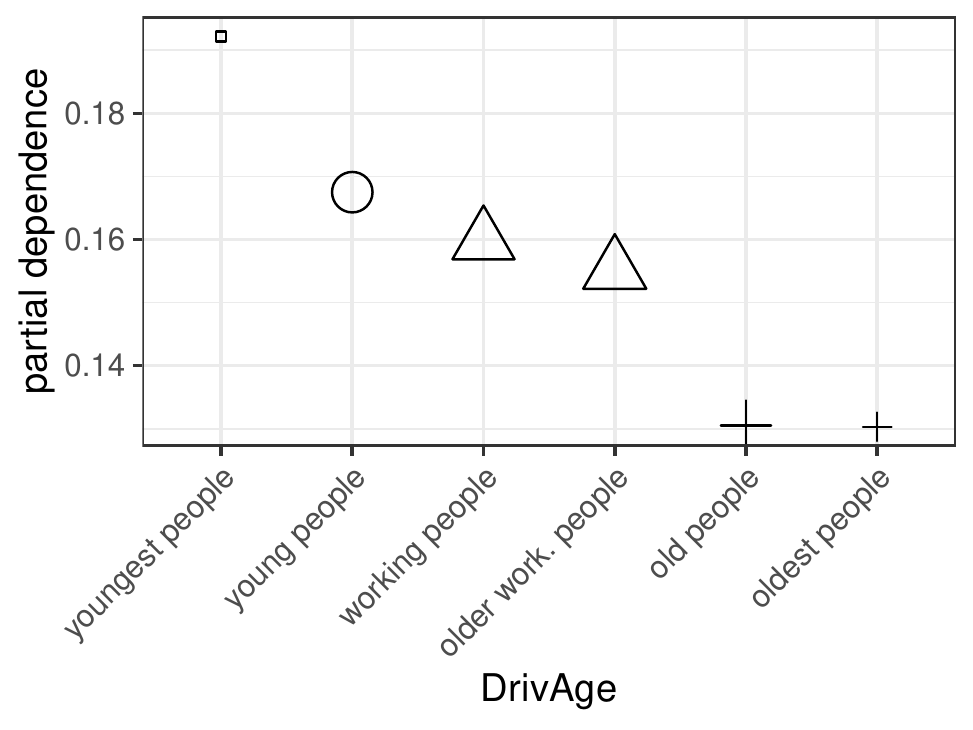}
		\label{ausprivauto_DrivAge}
	}
	\subfigure[\texttt{norauto}: driving distance limit]{
		\includegraphics[width=0.475\textwidth]{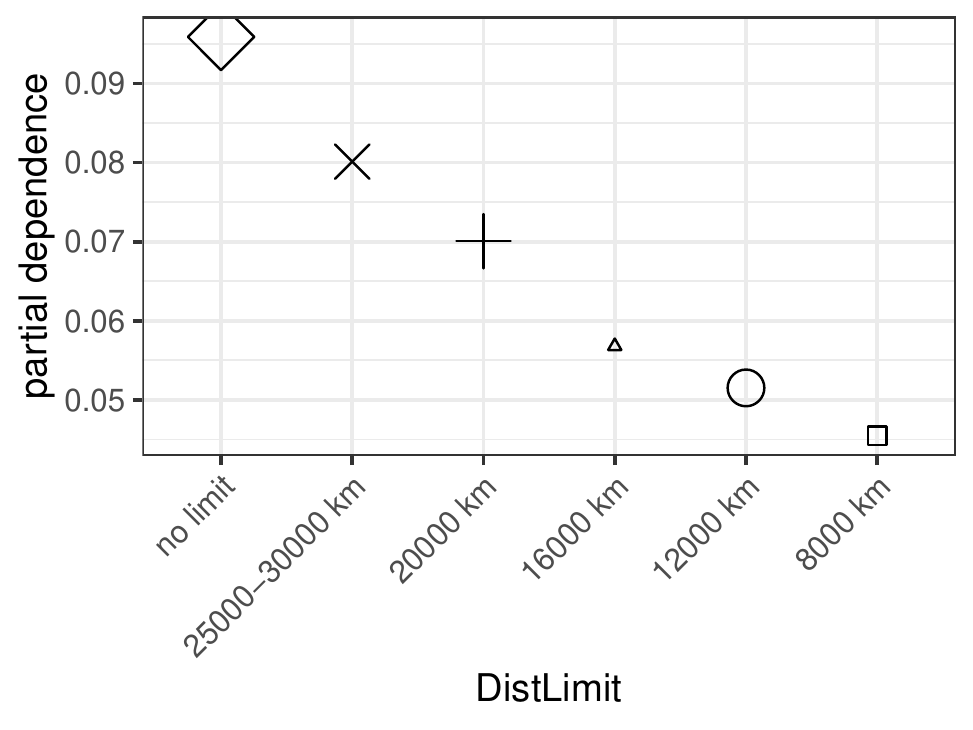}
		\label{norauto_DistLimit}
	}
	\subfigure[\texttt{pricingame}: make of the vehicle]{
		\includegraphics[width=0.95\textwidth]{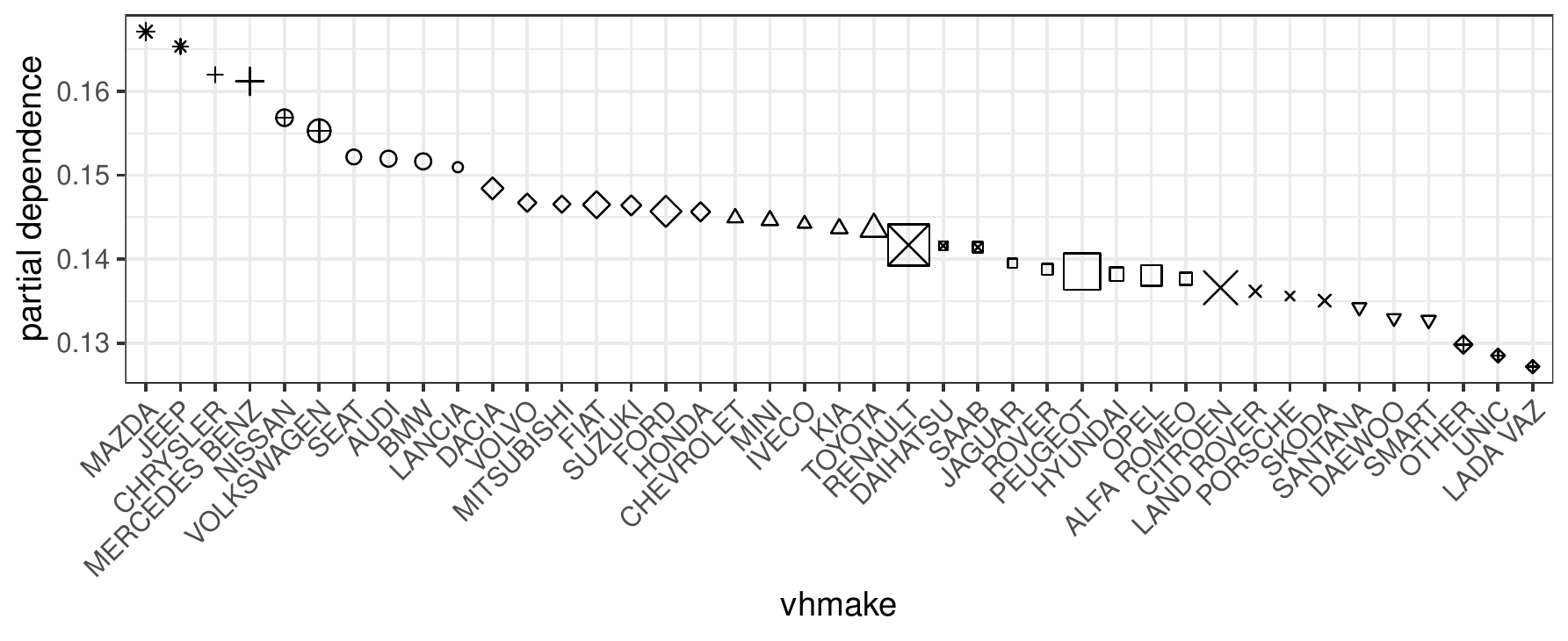}
		\label{pricingame_vhmake}
	}
	\caption{PD effect and the resulting segmentation for three categorical features. Groups are indicated by plotting characters, with size proportional to the observation count in the portfolio.}
	\label{segm_cat}
\end{figure}

Figure~\ref{segm_cat} displays the resulting segmentation for three categorical features. Groups are indicated by different plotting characters, with size proportional to the observation count in the portfolio. Figure~\ref{ausprivauto_DrivAge} shows that claim risk decreases with increasing age of the policyholder in the \texttt{ausprivauto} portfolio. Due to similar PD effects, both levels containing the oldest policyholders are grouped together as well as both levels containing the people of working age. This results in four age segments: youngest, young, working and older people. Figure~\ref{norauto_DistLimit} shows that claim risk decreases for a decreasing driving distance limit in the \texttt{norauto} portfolio. The PD effects are dissimilar enough not to be grouped together, so each level remains in a separate segment. Figure~\ref{pricingame_vhmake} shows the PD effects and resulting grouping for vehicle makes in the \texttt{norauto} portfolio. The 41 different makes are divided in 11 segments with \{Mazda, Jeep\} and \{Lada, Unic, Other\} as the most and least risky segments respectively. Categorical features with many levels are often hard to deal with in practice. Appendix~\ref{app_segm_geo} demonstrates how maidrr greatly reduces the complexity for geographical information in the \texttt{bemtpl} and \texttt{pricingame} portfolios.

\subsection{Evaluation of the GLM surrogate}
\label{eval}
This section evaluates the performance of the maidrr GLM surrogate based on three desiderata listed in \citet[Section 3.2]{Guidotti2018}: accuracy, fidelity and interpretability. Section~\ref{eval_acc} evaluates accuracy since generating accurate predictions is very important for a model to remain competitive and relevant in production. Section~\ref{eval_fid} evaluates fidelity as the extent to which the surrogate is able to mimic the behavior of a black box. Section~\ref{eval_int} evaluates interpretability because the surrogate should be comprehensible and easy to use in practice. We benchmark our GLM against two transparent surrogates: a decision tree (DT) and linear model (LM). Both are fit with the original data as features and the GBM predictions as target \citep{Molnar2020}. We restrict the maximum tree depth to four to keep the result comprehensible.

\subsubsection{Accuracy}
\label{eval_acc}
The goal of our maidrr GLM surrogate is to approximate a complex black box and replace it in the production pipeline. In order to justify this substitution, it is vital that the GLM results in accurate predictions with minimal accuracy loss compared to the black box. We measure prediction accuracy for all models via the Poisson deviance from Eq.~\eqref{poiss_dev}. With~$f_{\textrm{surro}}$ and~$f_{\textrm{gbm}}$ the surrogate and GBM prediction function, we assess the accuracy loss via percentage differences as follows:~$\Delta D^{\textrm{Poi}} = 100 \times \left(  D^{\textrm{Poi}} \{y, f_{\textrm{surro}}(\boldsymbol{x}) \} / D^{\textrm{Poi}} \{ y, f_{\textrm{gbm}}(\boldsymbol{x}) \} - 1 \right) $. 

Table~\ref{poi_diff} shows the Poisson percentage differences~$\Delta D^{\textrm{Poi}}$ for the GLM, LM and DT surrogates with respect to the GBM black box. Results are shown for each dataset separately and the last column contains the average over all datasets. The maidrr GLM attains the lowest accuracy loss and outperforms the benchmark surrogates on each dataset. The GLM's accuracy loss stays below \num{0.5}\% on four out of six datasets, with an average of \num{0.64}\% over all datasets. On average, the GLM is 3 and 7.5 times as accurate as the DT and LM surrogates.

\begin{table}[h!]
\centering
\caption{Poisson percentage differences $\Delta D^{\textrm{Poi}}$ for the different surrogate models.}
\begin{tabular}{lcccccc|c}
  \toprule
 & \texttt{ausprivauto} & \texttt{bemtpl} & \texttt{frempl} & \texttt{fremtpl} & \texttt{norauto} & \texttt{pricingame} & avg. \\ 
  \midrule
GLM & \bf 0.10 & \bf 0.49 & \bf 1.80 & \bf 0.92 & \bf 0.03 & \bf 0.48 & \bf 0.64 \\ 
  LM & 0.22 & 1.15 & 18.39 & 6.35 & 0.07 & 2.53 & 4.79 \\ 
  DT & 0.25 & 1.68 & 4.82 & 2.66 & 0.28 & 2.13 & 1.97 \\ 
   \bottomrule
\end{tabular}
\label{poi_diff}
\end{table}

\subsubsection{Fidelity}
\label{eval_fid}
This section investigates how closely the maidrr GLM mimics the behavior of the GBM black box by assessing how well the surrogates replicate the GBM's predictions.

The~$R^2$ measure represents the percentage of variance that the surrogate model is able to capture from the black box. With~$\mu_{\textrm{gbm}}$ the mean GBM prediction, the~$R^2 \in [0,1]$ is defined as follows:
\begin{equation*}
R^2 = 1 - \frac{\sum_{i=1}^n \left\lbrace  f_{\textrm{surro}}(\boldsymbol{x}_i) -  f_{\textrm{gbm}}(\boldsymbol{x}_i)\right\rbrace^2 }{\sum_{i=1}^n \left\lbrace  f_{\textrm{gbm}}(\boldsymbol{x}_i) -  \mu_{\textrm{gbm}}\right\rbrace^2}.
\label{rsquared_form}
\end{equation*}

Furthermore, we also compute Pearson's linear and Spearman's rank correlation coefficients~$\rho$ between the GBM and surrogate predictions. We average these coefficients to consolidate both types of correlation in one number, but the results below also hold for each coefficient separately.

Table~\ref{r_squared} shows the~$R^2$ for the GLM, LM and DT surrogates on each dataset separately and averaged over all datasets in the last column. The GLM ranks first in five datasets and second in \texttt{ausprivauto}. The GLM captures more than \num{90}\% of variance on four out of six datasets, with an average of \num{90}\% over all datasets. On average, the GLM captures an extra 12\% and 15\% of variance compared to the DT and LM surrogates.

\begin{table}[h!]
	\centering
	\caption{$R^2$ measure for the different surrogate models.}
	\begin{tabular}{lcccccc|c}
		\toprule
		& \texttt{ausprivauto} & \texttt{bemtpl} & \texttt{frempl} & \texttt{fremtpl} & \texttt{norauto} & \texttt{pricingame} & avg. \\ 
		\midrule
		GLM & 0.86 & \bf 0.94 & \bf 0.91 & \bf 0.78 & \bf 0.99 & \bf 0.93 & \bf 0.90 \\ 
  		LM & \bf 0.89 & 0.83 & 0.62 & 0.30 & 0.95 & 0.88 & 0.75 \\ 
  		DT & 0.75 & 0.74 & 0.88 & 0.75 & 0.84 & 0.76 & 0.78 \\ 
		\bottomrule
	\end{tabular}
	\label{r_squared}
\end{table}

Table~\ref{avg_rho} shows the averaged~$\rho$ for the GLM, LM and DT surrogates on each dataset separately and averaged over all datasets in the last column. The GLM ranks first in all datasets, thereby outperforming both benchmark surrogates. The correlation between the GBM and GLM is at least \num{95}\% on four out of six datasets, with an average of \num{95}\% over all datasets. On average, the GLM's correlation to the GBM is 12\% and 9\% higher compared to the DT and LM surrogates.

\begin{table}[h!]
	\centering
	\caption{Average correlation coefficient~$\rho$ for the different surrogate models.}
	\begin{tabular}{lcccccc|c}
		\toprule
		& \texttt{ausprivauto} & \texttt{bemtpl} & \texttt{frempl} & \texttt{fremtpl} & \texttt{norauto} & \texttt{pricingame} & avg. \\ 
		\midrule
		GLM & \bf 0.95 & \bf 0.97 & \bf 0.91 & \bf 0.92 & \bf 0.99 & \bf 0.97 & \bf 0.95 \\ 
  		LM & 0.95 & 0.93 & 0.74 & 0.60 & 0.98 & 0.95 & 0.86 \\ 
  		DT & 0.86 & 0.83 & 0.75 & 0.78 & 0.91 & 0.87 & 0.83 \\ 
		\bottomrule
	\end{tabular}
	\label{avg_rho}
\end{table}

In both Tables~\ref{r_squared} and~\ref{avg_rho}, the DT outperforms the LM on the \texttt{frempl} and \texttt{fremtpl} datasets while the LM outperforms the DT on the remaining four datasets. This is driven by the fact that the DT puts focus on interactions while the LM puts focus on marginal effects. Our maidrr GLM combines both marginal and interaction effects, resulting in better performance overall.

We conclude that our GLM constructed with maidrr outperforms the benchmark DT and LM surrogates when it comes to both prediction accuracy and mimicking the GBM's underlying behavior. Remember that the DT and LM are trained with the GBM's predictions as target. The maidrr procedure extracts knowledge from the GBM to perform smart feature engineering, but afterwards the GLM is fit to the original target. The observation that the GLM is better at mimicking the GBM compared to the benchmark surrogates is therefore especially interesting.

\subsubsection{Interpretability}
\label{eval_int}

\paragraph{Global interpretations}
A GLM is globally interpretable as the model coefficients, relating the features to the predictions, are easily observable. For example, the Poisson GLM with logarithmic link function to model the number of claims for the \texttt{norauto} dataset has the following structure and fitted coefficients:

\footnotesize
\begin{equation*}
\begin{split}
\ln \left(\mathbb{E}[nclaims] \right) = & -2.40 + 0.54 \, Male_{no} + 0.09 \, Young_{yes}  \\
											& - 0.76 \, DistLimit_{8000km} - 0.62 \, DistLimit_{12000km} \\
									 	 & - 0.51 \, DistLimit_{16000km} - 0.33 \, DistLimit_{20000km} - 0.20 \, DistLimit_{30000km} \\
									 	 & - 0.17 \, GeoRegion_{Low- \,\,\& \,\, Low+} - 0.05 \, GeoRegion_{Med-} + 0.23 \, GeoRegion_{High+} \\
									 	 & - 0.08 \, DistLimit\_\_GeoRegion_{8000/12000/16000km\_\_High+ \,\, \& \,\, nolimit\_\_Low-/Low+/Med-}
\end{split}
\end{equation*}
\normalsize
where $Male_{yes}$, $Young_{no}$, $DistLimit_{nolimit}$ and $GeoRegion_{Med+ \,\, \& \,\, High-}$ are the reference levels captured by the intercept. These references are the levels which contain the highest number of policyholders such that the intercept models the claim frequency of an ``average'' policyholder. Taking the inverse link function, namely the exponential, on both sides results in a multiplicative GLM prediction function with the following global interpretations:
\begin{itemize}
	\setlength\itemsep{1pt}
	\item The predicted claim frequency for an older male policyholder without a driving distance limit and living in the Med+ or High- geographical region equals 0.09 or $\exp(-2.40)$.
	\item Predictions are 72\% higher for female policyholders compared to males as $\exp(0.54) = 1.72$. \textit{Note: In 2012, the EU put forward rules on gender-neutral pricing in the insurance industry such that gender is no longer allowed as a rating factor in a commercial tariff.}
	\item As $\exp(0.09) = 1.09$, predictions are 9\% higher for young compared to old policyholders.
	\item For policyholders with a driving distance limit of 8, 12, 16, 20 and 30 thousand kilometers, predictions respectively amount to 47\%, 54\%, 60\%, 72\% and 82\% of those for someone without a limit. There is a clear increasing trend of claim risk with the distance limit.
	\item Predictions for policyholders living in the Low or Med- geographical regions amount to respectively 84\% and 95\% of those for the Med+/High- regions, whereas predictions increase with 26\% for those in the High+ region.
	\item The interaction between the distance limit and geographical region results in a negative correction for policyholders with the most risky level of one of the features and a low-risk level of the other. As $\exp(0.08) = 0.92$, predictions are reduced by 8\% for policyholders living in the High+ region with a maximal distance limit of \num{16000} kilometer and for those with no distance limit which live in the Low-, Low+ or Med- region. 
\end{itemize}
Our maidrr procedure outputs a GLM with all features in a categorical format such that the full working regime of the GLM can be summarized in a decision table, see Appendix~\ref{app_glm_table}. In practice, such a tabular model is easy to represent and maintain in a spreadsheet with responsive filters. Decision tables are very comprehensible for human users and outperform both trees and rules in accuracy, response time, answer confidence and ease of use \citep{Huysmans2011}.

\paragraph{Local interpretations}
We now turn to explaining individual predictions for the three artificial instances in the \texttt{bemtpl} dataset listed in Table~\ref{instances}. Based on the GBM and GLM predictions, these instances represent a high/medium/low risk profile. We want to assess how the features influence the riskiness of each individual. Feature contributions in a GLM can be extracted via the fitted coefficients, as implemented in \texttt{maidrr::explain} \citep{maidrr}. For comparison purposes we use \citet{Shapley1953} values to explain the GBM predictions, with the efficient implementation of \citet{Strumbelj2010,Strumbelj2014} available in the \textsf{R} package \texttt{iml} \citep{Molnar2018}.

\begin{table}[h!]
	\centering
	\scriptsize
	\caption{Artificial instances in the \texttt{bemtpl} portfolio for which we explain the individual predictions.}
	\begin{tabular}{lccc}
		\toprule
		& high risk &medium risk & low risk\\
		\midrule
		\texttt{bm}&7&4&1\\
		\texttt{postcode}&11&91&55\\
		\texttt{ageph}&27&45&66\\
		\texttt{power}&96&66&26\\
		\texttt{fuel}&diesel&gasoline&gasoline\\
		\texttt{agec}&1&8&15\\
		\texttt{coverage}&TPL&TPL+&TPL++\\
		\texttt{fleet}&yes&no&no\\
		\texttt{sex}&female&male&male\\
		\texttt{use}&private&work&work\\
		\midrule
		GBM&\num{0.2847}&\num{0.1398}&\num{0.0502}\\
		GLM&\num{0.3861}&\num{0.1231}&\num{0.0413}\\
		\bottomrule
	\end{tabular}
	\label{instances}
\end{table}

\begin{figure}[h!]
	\centering
	\subfigure[GBM: high risk]{
		\includegraphics[width=0.31\textwidth]{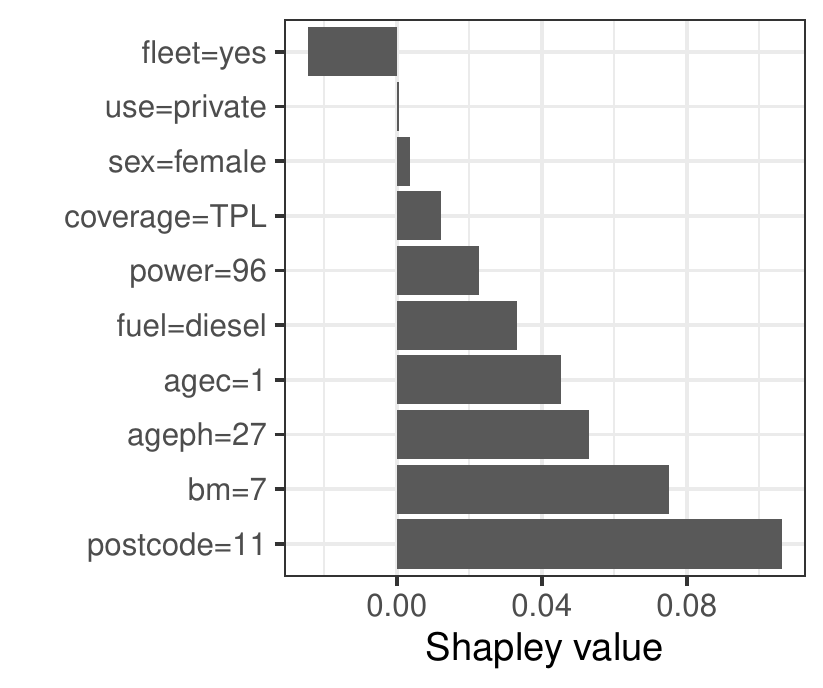}
		\label{shap1}
	}
	\subfigure[GBM: medium risk]{
		\includegraphics[width=0.31\textwidth]{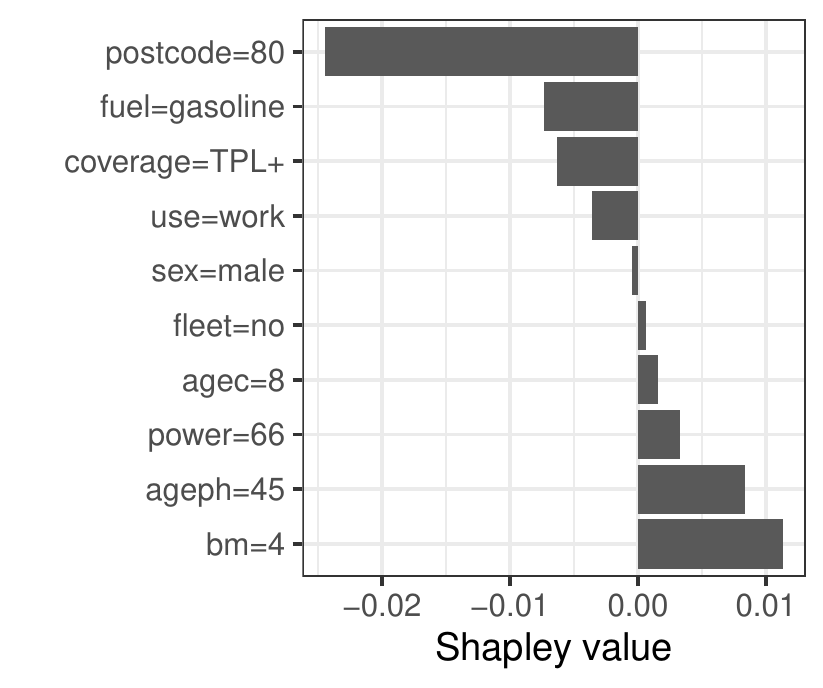}
		\label{shap2}
	}
	\subfigure[GBM: low risk]{
		\includegraphics[width=0.31\textwidth]{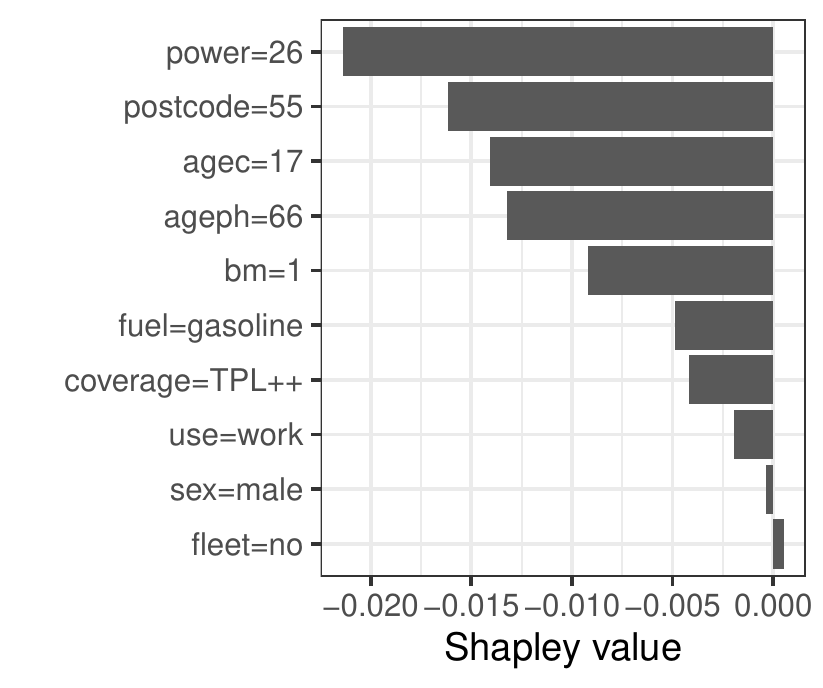}
		\label{shap3}
	}
	\subfigure[GLM: high risk]{
		\includegraphics[width=0.31\textwidth]{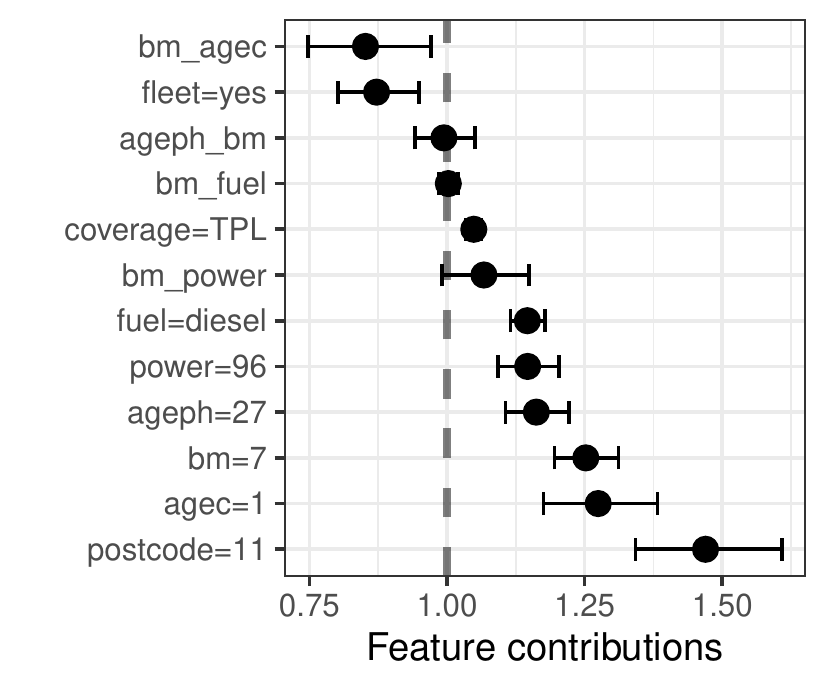}
		\label{iglm1}
	}
	\subfigure[GLM: medium risk]{
		\includegraphics[width=0.31\textwidth]{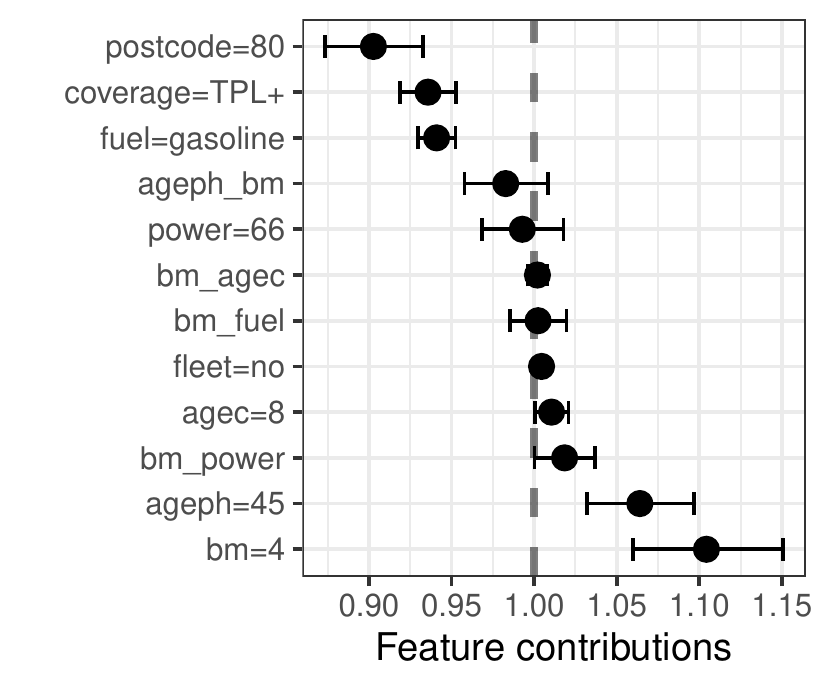}
		\label{iglm2}
	}
	\subfigure[GLM: low risk]{
		\includegraphics[width=0.31\textwidth]{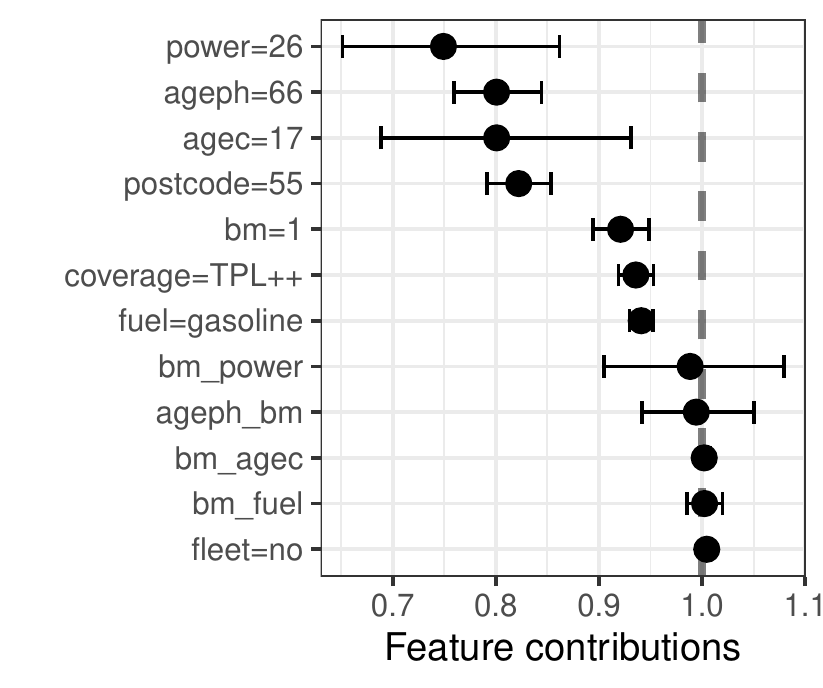}
		\label{iglm3}
	}
	\caption{Explanations for the high (left), medium (middle) and low (right) risk instance predictions from Table~\ref{instances} in the GBM via Shapley values (top) and the GLM via $\beta$ coefficients (bottom).}
	\label{exp_gbm_glm}
\end{figure}

Figures \ref{shap1}, \ref{shap2} and \ref{shap3} show the Shapley values for the GBM prediction of each instance. The sum of these values equals the difference between the instance prediction, shown in Table~\ref{instances}, and the average GBM prediction of \num{0.1417}. The presence of mainly positive/negative Shapley values in Figure~\ref{shap1}/\ref{shap3} thus represents a high/low risk profile respectively. Figures~\ref{iglm1},~\ref{iglm2} and~\ref{iglm3} show the GLM's feature contributions on the response scale after taking the inverse link function, namely $\exp(\beta_j)$ for feature $x_j$ in our Poisson GLMs with log link. The contributions are multiplicative with respect to the baseline prediction of \num{0.13}, as captured by the intercept, and the gray dashed line indicates the point of ``no contribution'' at $\exp(0)$. Furthermore, the GLM allows to split the contributions over marginal effects and interactions with other features, while 95\% confidence intervals indicate the uncertainty associated with each contribution.

The GBM and GLM explanations are very similar. For example, Figures~\ref{shap1} and~\ref{iglm1} attribute this profile's high risk to a residence in Brussels, young age, high bonus-malus level and driving a new high-powered diesel vehicle. The interaction between the bonus-malus level and age of the vehicle puts a negative correction on both positive marginal effects in the GLM, while the other interactions have limited impact on the prediction. The GLMs show no contribution from gender as this feature is not selected by maidrr, while it has negligibly small Shapley values in all cases. An insurance rate is determined by the product of claim frequency and severity, such that the contributions can be directly interpreted as a percentage premium/discount on the price. Living in Brussels increases the baseline frequency, and thus the price, by almost 50\% in the technical analysis for this dataset. One can assess the fairness of this penalty, possibly followed by a manual adjustment to intervene in the decision-making process via expert judgment.

\section{Conclusions}
\label{concl}
Decision-making algorithms in business practice can become highly complex in order to gain a competitive advantage. However, transparency is a key requirement for any high-stakes decision or for companies active in strictly regulated industries. To balance accuracy and explainability, we present maidrr: a procedure to develop a Model-Agnostic Interpretable Data-driven suRRogate for a complex system. The paper is accompanied by an \textsf{R} package in which the procedure is implemented \citep{maidrr}. We apply maidrr to six real-life general insurance portfolios for claim frequency prediction, with insurance pricing as an example of a high-stakes decision in a strictly regulated industry. We thereby also put focus on a highly relevant count regression problem, which is not often dealt with in classical machine learning. Our maidrr procedure results in a surrogate GLM which closely approximates the performance of a black box GBM in terms of accuracy and fidelity, while outperforming two benchmark surrogates. This allows to substitute the GLM in the production pipeline with minimal performance loss. In the process, maidrr automatically performs feature selection and segmentation, providing a possibly useful by-product for customer or market segmentation applications.

Both global and local interpretations are easily extracted from our maidrr GLM. Explanations only depend on the fitted coefficients, which are easily observable and presentable on the response scale. This representation boosts the ability to understand the feature contributions on the scale of interest and allows for manual intervention when deploying the model in practice. This gives some important advantages to maidrr with respect to the following XAI goals \citep[see][Table~1]{Arrieta2020}. 1)~\emph{Trustworthiness:} a GLM with only categorical features always acts as intended since all the possible working regimes can be listed in a decision table of fixed size. 2)~\emph{Accessibility/Interactivity:} manual post-processing of the model becomes very easy and intuitive by tweaking the GLM coefficients. This allows users to intervene and be more involved in the development and improvement of the model. 3)~\emph{Fairness:} the clear influence of each feature allows for an ethical analysis of the model, which becomes especially important for high-stakes decisions which influence people's lives. In our insurance setting, it is important that every policyholder receives a fair insurance quote. The direct interpretation of the feature contributions as a penalty/discount to the baseline tariff further serves this cause. 4)~\emph{Confidence:} the uncertainty of the contributions is quantifiable via confidence intervals such that the model's robustness, stability and reliability can be assessed. 5)~\emph{Informativeness:} contributions are split across marginal effects and interactions of features, thereby increasing the amount of information available to the user on the underlying decision of the model. 

Our maidrr procedure combines the inherent interpretability of a GLM with the accurate approximation of a sophisticated black box. We therefore believe that maidrr can serve as a useful tool in any situation where a competitive, yet transparent model is needed.

\section*{Funding}
This research is supported by the Research Foundation Flanders [SB grant 1S06018N] and by the Natural Sciences and Engineering Research Council of Canada [RGPIN-2019-04190]. Furthermore, Katrien Antonio acknowledges financial support from the Ageas Research Chair at KU Leuven and from KU Leuven's research council [COMPACT C24/15/001].

\newpage

\appendix

\section{PD and ALE for correlated features}
\label{app_corr_feat}
\begin{figure}[h!]
	\centering
	\subfigure[correlation matrix]{
		\includegraphics[width=0.4\textwidth]{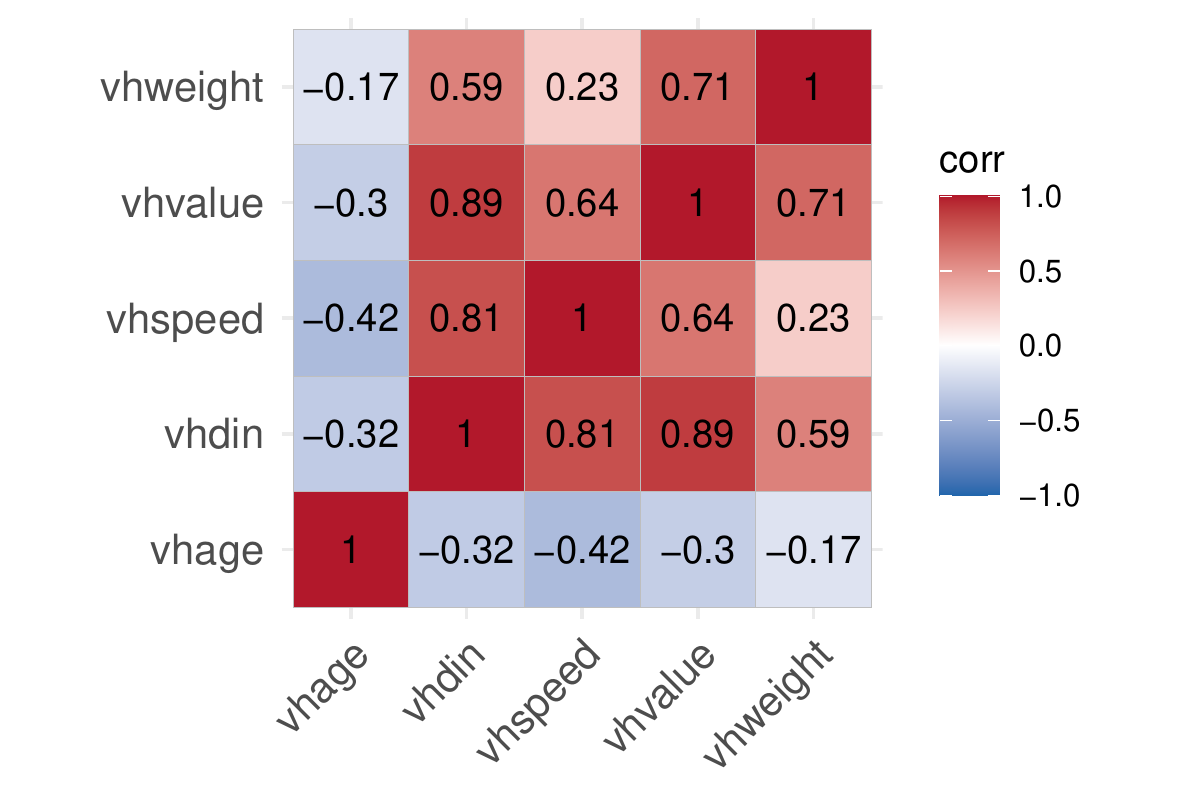}
		\label{corr_mat}
	}
	\subfigure[age in years]{
		\includegraphics[width=0.4\textwidth]{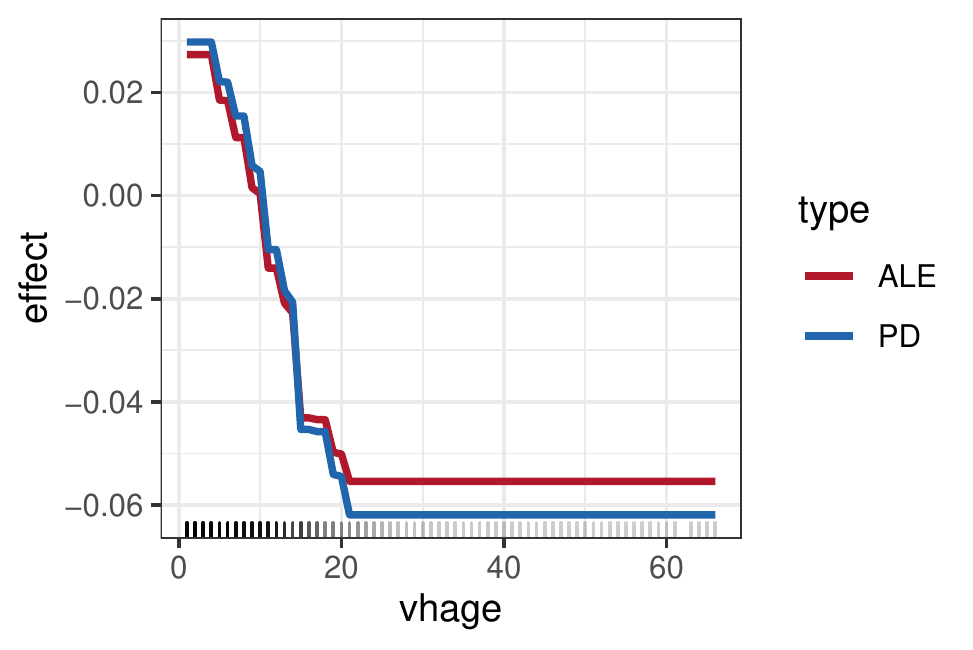}
		\label{corr_age}
	}
	\subfigure[motor power in hp]{
		\includegraphics[width=0.4\textwidth]{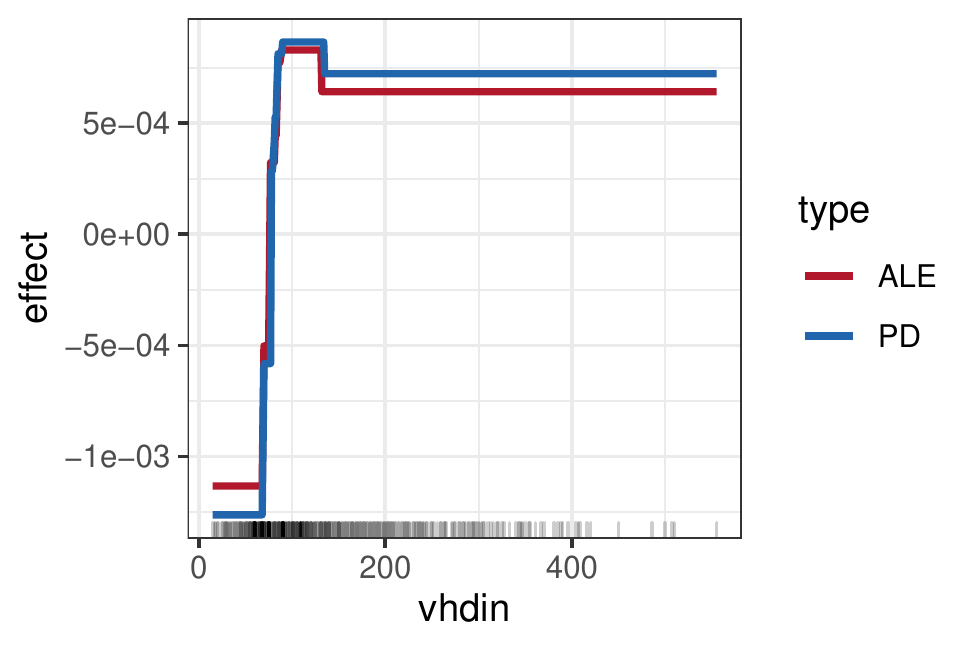}
		\label{corr_din}
	}
	\subfigure[maximum speed in km/h]{
		\includegraphics[width=0.4\textwidth]{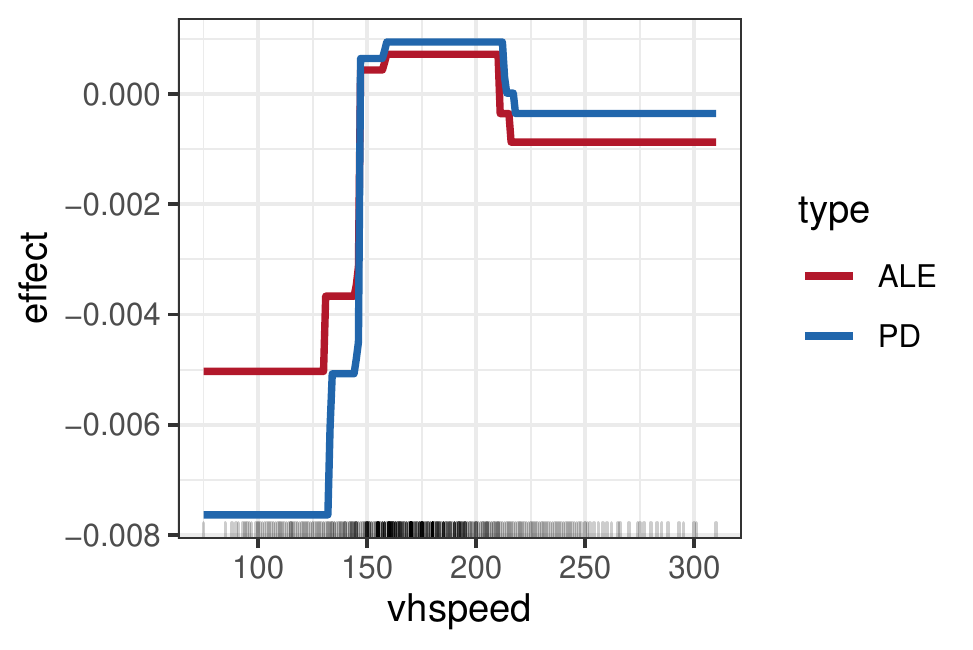}
		\label{corr_speed}
	}
	\subfigure[value in euros]{
		\includegraphics[width=0.4\textwidth]{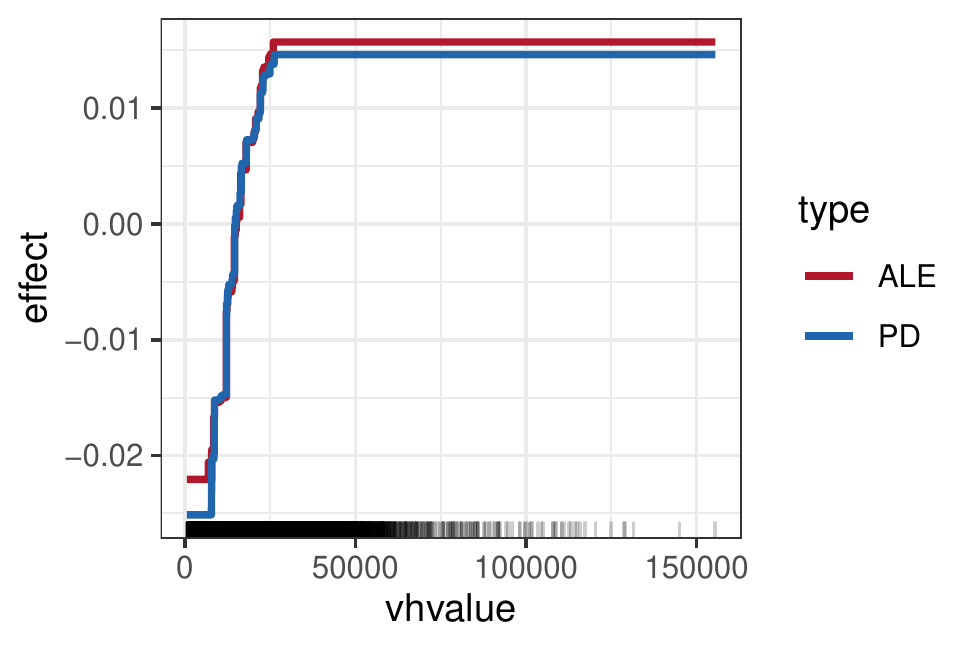}
		\label{corr_value}
	}
	\subfigure[weight in kg]{
		\includegraphics[width=0.4\textwidth]{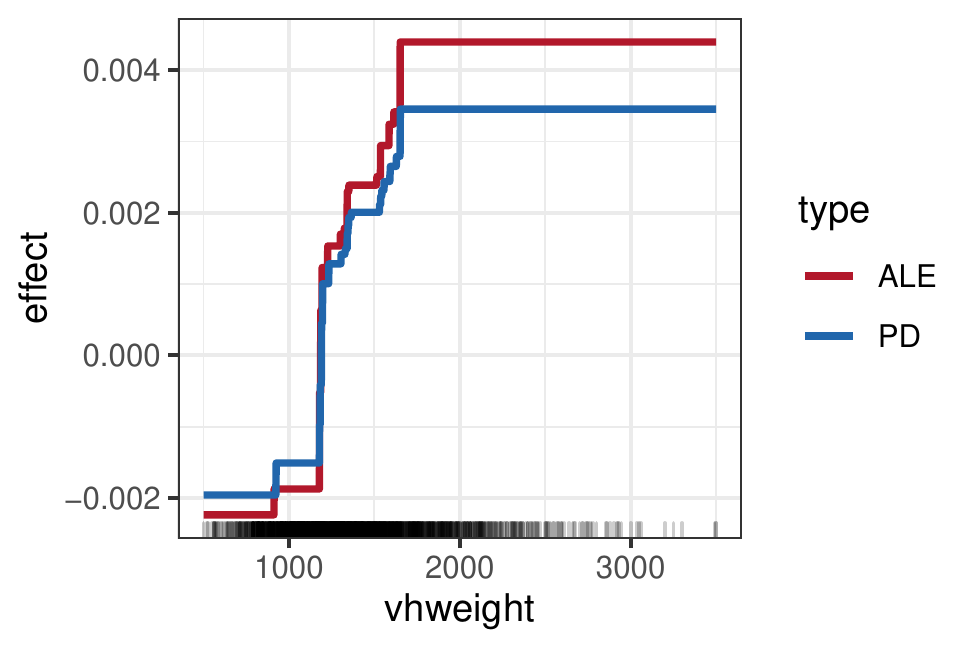}
		\label{corr_weight}
	}
	\caption{Comparison of PD and ALE for correlated vehicle characteristics in the \texttt{pricingame} dataset.}
	\label{corr_feat}
\end{figure}

Figure~\ref{corr_feat} compares the PD and ALE for several vehicle characteristics in the \texttt{pricingame} dataset, namely the weight, value, maximum speed, horsepower and age. Figure~\ref{corr_mat} shows that the vehicle age is negatively correlated with the other characteristics while there is a strong positive correlation between the weight, value, maximum speed and horsepower. Figures~\ref{corr_age}, \ref{corr_din}, \ref{corr_speed}, \ref{corr_value} and \ref{corr_weight} show the centered PD (in blue) and ALE (in red) for all the vehicle features. Both effects are very similar for each of the features, especially in the ranges with high observation counts as indicated by the black rugs on the x-axis. We observe some vertical shifts between the PD and ALE in feature ranges with low observation counts. However, these vertical shifts are not a problem for our maidrr procedure as we only use these effects to perform the feature grouping. Furthermore, observation counts are taken into account as weights in the penalized loss function of Eq.~\eqref{pen_loss}, further reducing the impact of these shifts on the obtained segmentation. This justifies the use of PD effects for grouping, even when dealing with correlated features.

\section{GLM formulation}
\label{app_glm_form}
A GLM allows any distribution from the exponential family for the target of interest $y$. This includes, among others, the normal, Bernoulli, Poisson and gamma distributions, making GLMs a versatile modeling tool. Denoting by $g(\cdot)$ the link function, the structure of a GLM with all features $\boldsymbol{x}$ in a categorical format is as follows:
\begin{equation*}
g(\mathbb{E}[y]) = \boldsymbol{\ell}^\top\boldsymbol{\beta} = \beta_0 + \sum_{j=1}^{d} \beta_j \ell_{j}.
\end{equation*}
The $d+1$ dimensional vector $\boldsymbol{\ell}$ contains a 1 for the intercept~$\beta_0$ together with $d$ dummy variables~$\ell_j \in \{0,1\}$. A categorical feature~$x$ with $m$ levels contains a reference level which is captured by the intercept. The other $m-1$ levels are coded via dummy variables to model the differences between those levels and the reference level, captured by the coefficients $\beta_j$. 

\section{GLM in a tabular format}
\label{app_glm_table}
Table~\ref{glm_table} shows part of the decision table for the \texttt{norauto} dataset, with the four lowest and highest predictions indicated in italics and bold respectively. The three other parts for \emph{Male = Yes \& Young = No} and \emph{Male = No \& Young = Yes/No} are not shown for space reasons. 

\begin{table}[h!]
	\centering
	\caption{Part of the GLM predictions in a decision table for the \texttt{norauto} dataset.}
	\footnotesize
	\begin{tabular}{lccccr}
		\toprule
		& Male & Young & DistLimit & GeoRegion & GLM prediction (\%) \\
		\midrule
		1 & Yes & Yes & 8000 km & Low- \& Low+ & \textit{3.88} \\ 
  2 & Yes & Yes & 8000 km & Medium- & \textit{4.41} \\ 
  3 & Yes & Yes & 8000 km & Medium+ \& High- & \textit{4.62} \\ 
  4 & Yes & Yes & 8000 km & High+ & 5.36 \\ 
  5 & Yes & Yes & 12000 km & Low- \& Low+ & \textit{4.47} \\ 
  6 & Yes & Yes & 12000 km & Medium- & 5.08 \\ 
  7 & Yes & Yes & 12000 km & Medium+ \& High- & 5.32 \\ 
  8 & Yes & Yes & 12000 km & High+ & 6.17 \\ 
  9 & Yes & Yes & 16000 km & Low- \& Low+ & 4.99 \\ 
  10 & Yes & Yes & 16000 km & Medium- & 5.67 \\ 
  11 & Yes & Yes & 16000 km & Medium+ \& High- & 5.94 \\ 
  12 & Yes & Yes & 16000 km & High+ & 6.89 \\ 
  13 & Yes & Yes & 20000 km & Low- \& Low+ & 5.94 \\ 
  14 & Yes & Yes & 20000 km & Medium- & 6.75 \\ 
  15 & Yes & Yes & 20000 km & Medium+ \& High- & 7.07 \\ 
  16 & Yes & Yes & 20000 km & High+ & \bf 8.92 \\ 
  17 & Yes & Yes & 30000 km & Low- \& Low+ & 6.78 \\ 
  18 & Yes & Yes & 30000 km & Medium- & 7.70 \\ 
  19 & Yes & Yes & 30000 km & Medium+ \& High- & 8.07 \\ 
  20 & Yes & Yes & 30000 km & High+ & \bf 10.18 \\ 
  21 & Yes & Yes & no limit & Low- \& Low+ & 7.63 \\ 
  22 & Yes & Yes & no limit & Medium- &  8.67 \\ 
  23 & Yes & Yes & no limit & Medium+ \& High- & \bf 9.87 \\ 
  24 & Yes & Yes & no limit & High+ & \bf 12.45 \\ 
		\bottomrule
	\end{tabular}
	\label{glm_table}
\end{table}

\section{Geographical segmentation}
\label{app_segm_geo}
\begin{figure}[htbp!]
	\centering
	\subfigure[\texttt{bemtpl}: postal code]{
		\includegraphics[width=0.62\textwidth]{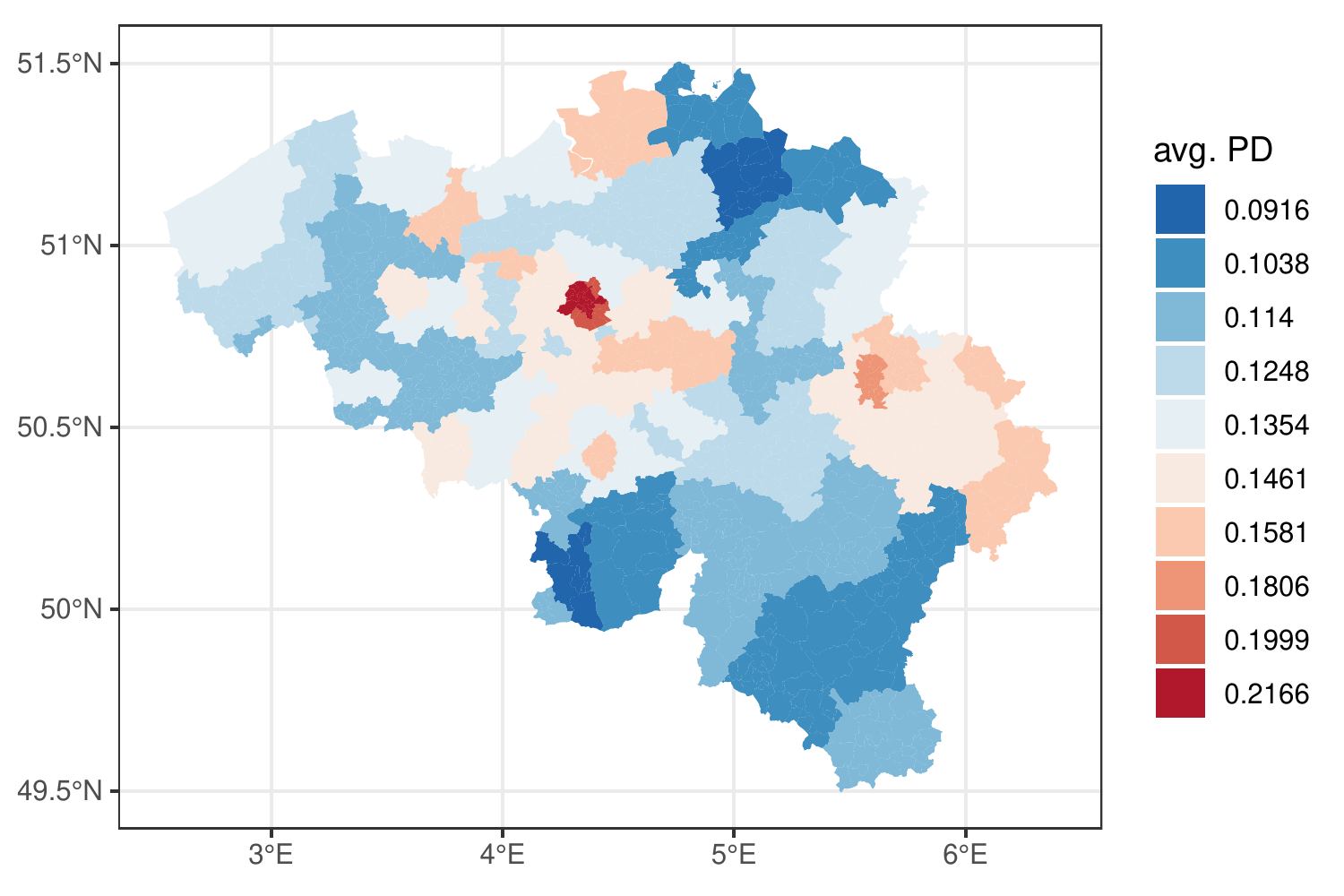}
		\label{bemtpl_postcode}
	}
	\subfigure[\texttt{pricingame}: INSEE department code]{
		\includegraphics[width=0.68\textwidth]{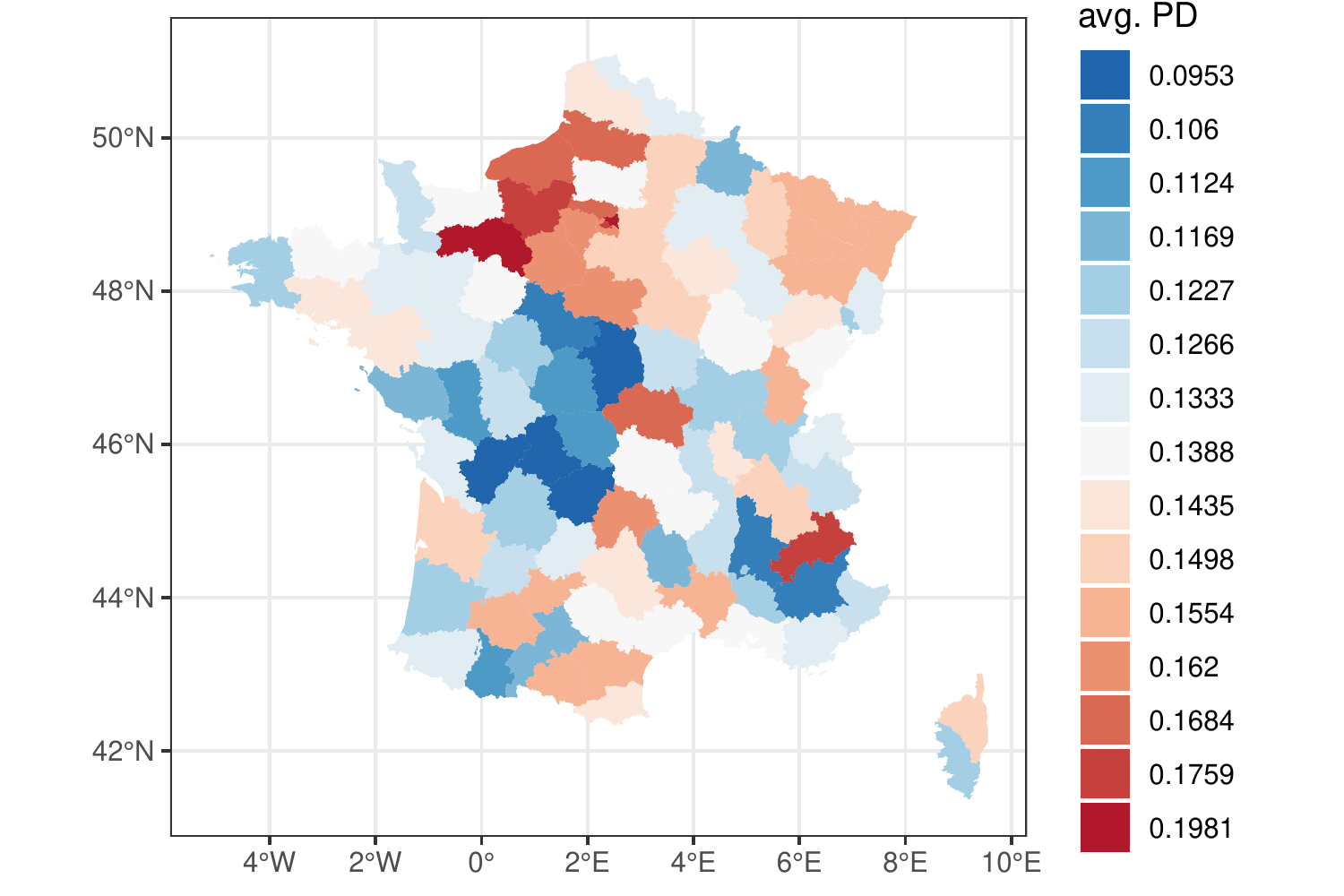}
		\label{pricingame_polinseecode}
	}
	\caption{Average PD effect for geographical regions where groups are indicated by colors.}
	\label{segm_geo}
\end{figure}

Figure~\ref{segm_geo} shows the average PD effect for geographical regions where groups are indicated by colors. Figure~\ref{bemtpl_postcode} shows the postal code areas on the map of Belgium with the initial \num{80} regions from the \texttt{bemtpl} portfolio segmented in \num{10} clusters. The capital Brussels in the center of Belgium (red colored), together with other big cities (orange colored), are risky due to heavy traffic in those densely populated areas. The rural regions in the northeast and south of Belgium are less risky. Figure~\ref{pricingame_polinseecode} shows the INSEE department code areas on the map of France with the initial \num{96} regions from the \texttt{pricingame} portfolio segmented in \num{15} clusters. The capital Paris and surrounding departments in the north of France (red/orange colored) are high-risk areas.

\bibliographystyle{plainnat} 
\small
\bibliography{bibfile}

\begin{thebibliography}{44}
\providecommand{\natexlab}[1]{#1}
\providecommand{\url}[1]{\texttt{#1}}
\expandafter\ifx\csname urlstyle\endcsname\relax
  \providecommand{\doi}[1]{doi: #1}\else
  \providecommand{\doi}{doi: \begingroup \urlstyle{rm}\Url}\fi

\bibitem[Ahmad et~al.(2018)Ahmad, Eckert, and Teredesai]{Ahmad2018}
M.~A. Ahmad, C.~Eckert, and A.~Teredesai.
\newblock Interpretable machine learning in healthcare.
\newblock In \emph{Proceedings of the 2018 ACM international conference on
  bioinformatics, computational biology, and health informatics}, pages
  559--560, 2018.

\bibitem[Apley and Zhu(2019)]{Apley2016}
D.~W. Apley and J.~Zhu.
\newblock Visualizing the effects of predictor variables in black box
  supervised learning models.
\newblock \emph{arXiv preprint arXiv:1612.08468}, 2019.

\bibitem[Arrieta et~al.(2020)Arrieta, D{\'\i}az-Rodr{\'\i}guez, Del~Ser,
  Bennetot, Tabik, Barbado, Garc{\'\i}a, Gil-L{\'o}pez, Molina, Benjamins,
  Chatila, and Herrera]{Arrieta2020}
A.~B. Arrieta, N.~D{\'\i}az-Rodr{\'\i}guez, J.~Del~Ser, A.~Bennetot, S.~Tabik,
  A.~Barbado, S.~Garc{\'\i}a, S.~Gil-L{\'o}pez, D.~Molina, R.~Benjamins,
  R.~Chatila, and F.~Herrera.
\newblock Explainable {A}rtificial {I}ntelligence ({XAI}): Concepts,
  taxonomies, opportunities and challenges toward responsible {AI}.
\newblock \emph{Information Fusion}, 58:\penalty0 82--115, 2020.

\bibitem[Ba and Caruana(2014)]{Ba2014}
J.~Ba and R.~Caruana.
\newblock Do deep nets really need to be deep?
\newblock In \emph{Advances in Neural Information Processing Systems 27}, pages
  2654--2662, 2014.

\bibitem[Biecek(2018)]{Biecek2018}
P.~Biecek.
\newblock {DALEX}: explainers for complex predictive models in {R}.
\newblock \emph{The Journal of Machine Learning Research}, 19\penalty0
  (1):\penalty0 3245--3249, 2018.

\bibitem[Bracke et~al.(2019)Bracke, Datta, Jung, and Sen]{Bracke2019}
P.~Bracke, A.~Datta, C.~Jung, and S.~Sen.
\newblock \emph{Machine learning explainability in finance: an application to
  default risk analysis}.
\newblock Bank of England Working Paper No. 816, 2019.

\bibitem[Bucil\u{a} et~al.(2006)Bucil\u{a}, Caruana, and
  Niculescu-Mizil]{Bucila2006}
C.~Bucil\u{a}, R.~Caruana, and A.~Niculescu-Mizil.
\newblock Model compression.
\newblock In \emph{Proceedings of the 12th ACM SIGKDD international conference
  on Knowledge discovery and data mining}, pages 535--541, 2006.

\bibitem[Doran et~al.(2017)Doran, Schulz, and Besold]{Doran2017}
D.~Doran, S.~Schulz, and T.~R. Besold.
\newblock What does explainable {AI} really mean? {A} new conceptualization of
  perspectives.
\newblock \emph{arXiv preprint arXiv:1710.00794}, 2017.

\bibitem[Dutang and Charpentier(2019)]{CASdatasets}
C.~Dutang and A.~Charpentier.
\newblock \emph{CASdatasets: Insurance datasets}, 2019.
\newblock URL \url{http://cas.uqam.ca}.
\newblock R package version 1.0.10.

\bibitem[ECOA(1974)]{ECOA}
ECOA.
\newblock {U.S. C}ode {T}itle 15. {C}ommerce and {T}rade.
\newblock \emph{Chapter 41. Consumer Credit Protection}, Subchapter IV. Equal
  Credit Opportunity (Section 1691), 1974.

\bibitem[Friedman(2001)]{Friedman2001}
J.~H. Friedman.
\newblock Greedy function approximation: a gradient boosting machine.
\newblock \emph{Annals of Statistics}, 29\penalty0 (5):\penalty0 1189--1232,
  2001.

\bibitem[Friedman(2002)]{Friedman2002}
J.~H. Friedman.
\newblock Stochastic gradient boosting.
\newblock \emph{Computational Statistics \& Data Analysis}, 38\penalty0
  (4):\penalty0 367--378, 2002.

\bibitem[Friedman and Popescu(2008)]{Friedman2008}
J.~H. Friedman and B.~E. Popescu.
\newblock Predictive learning via rule ensembles.
\newblock \emph{The Annals of Applied Statistics}, 2\penalty0 (3):\penalty0
  916--954, 2008.

\bibitem[Gade et~al.(2019)Gade, Geyik, Kenthapadi, Mithal, and Taly]{Gade2019}
K.~Gade, S.~C. Geyik, K.~Kenthapadi, V.~Mithal, and A.~Taly.
\newblock Explainable {AI} in industry.
\newblock In \emph{Proceedings of the 25th ACM SIGKDD International Conference
  on Knowledge Discovery \& Data Mining}, pages 3203--3204, 2019.

\bibitem[GDPR(2016)]{GDPR}
GDPR.
\newblock {R}egulation ({EU}) 2016/679 of the {E}uropean {P}arliament and of
  the {C}ouncil of 27 {A}pril 2016 on the protection of natural persons with
  regard to the processing of personal data and on the free movement of such
  data.
\newblock \emph{O.J. (L 119)}, 1:\penalty0 1--88, 2016.

\bibitem[Greenwell et~al.(2019)Greenwell, Boehmke, Cunningham, and
  Developers]{gbm}
B.~Greenwell, B.~Boehmke, J.~Cunningham, and GBM Developers.
\newblock \emph{gbm: Generalized Boosted Regression Models}, 2019.
\newblock URL \url{https://cran.r-project.org/package=gbm}.
\newblock R package version 2.1.6.

\bibitem[Guidotti et~al.(2018)Guidotti, Monreale, Ruggieri, Turini, Giannotti,
  and Pedreschi]{Guidotti2018}
R.~Guidotti, A.~Monreale, S.~Ruggieri, F.~Turini, F.~Giannotti, and
  D.~Pedreschi.
\newblock A survey of methods for explaining black box models.
\newblock \emph{ACM Computing Surveys}, 51\penalty0 (5):\penalty0 1--42, 2018.

\bibitem[Gunning(2017)]{Gunning2017}
D.~Gunning.
\newblock Explainable {A}rtificial {I}ntelligence ({XAI}).
\newblock \emph{Defense Advanced Research Projects Agency (DARPA)}, Tech. rep.,
  2017.

\bibitem[Hall et~al.(2017)Hall, Gill, Kurka, and Phan]{Hall2017}
P.~Hall, N.~Gill, M.~Kurka, and W.~Phan.
\newblock \emph{Machine learning interpretability with {H}2{O} {D}riverless
  {AI}}.
\newblock H2O.ai, 2017.
\newblock URL
  \url{http://docs.h2o.ai/driverless-ai/latest-stable/docs/booklets/MLIBooklet.pdf}.

\bibitem[Hastie et~al.(2009)Hastie, Tibshirani, and Friedman]{Hastie2009}
T.~Hastie, R.~Tibshirani, and J.~Friedman.
\newblock \emph{The elements of statistical learning: data mining, inference,
  and prediction}.
\newblock Springer, New York, 2009.

\bibitem[Henckaerts(2020)]{maidrr}
R.~Henckaerts.
\newblock \emph{maidrr: Model-Agnostic Interpretable Data-driven suRRogate},
  2020.
\newblock URL \url{https://github.com/henckr/maidrr}.
\newblock R package version 1.0.0.

\bibitem[Henckaerts et~al.(2020)Henckaerts, C{\^o}t{\'e}, Antonio, and
  Verbelen]{Henckaerts2020}
R.~Henckaerts, M.~P. C{\^o}t{\'e}, K.~Antonio, and R.~Verbelen.
\newblock Boosting insights in insurance tariff plans with tree-based machine
  learning methods.
\newblock \emph{North American Actuarial Journal}, pages 1--31, 2020.

\bibitem[Hinton et~al.(2015)Hinton, Vinyals, and Dean]{Hinton2015}
G.~Hinton, O.~Vinyals, and J.~Dean.
\newblock Distilling the knowledge in a neural network.
\newblock \emph{arXiv preprint arXiv:1503.02531}, 2015.

\bibitem[Hrnjica and Softic(2020)]{Hrnjica2020}
B.~Hrnjica and S.~Softic.
\newblock Explainable {AI} in manufacturing: A predictive maintenance case
  study.
\newblock In \emph{Advances in Production Management Systems. Towards Smart and
  Digital Manufacturing}, pages 66--73. Springer, 2020.

\bibitem[Hu et~al.(2020)Hu, Chen, Nair, and Sudjianto]{Hu2020}
L.~Hu, J.~Chen, V.~N. Nair, and A.~Sudjianto.
\newblock Surrogate locally-interpretable models with supervised machine
  learning algorithms.
\newblock \emph{arXiv preprint arXiv:2007.14528}, 2020.

\bibitem[Huysmans et~al.(2011)Huysmans, Dejaeger, Mues, Vanthienen, and
  Baesens]{Huysmans2011}
J.~Huysmans, K.~Dejaeger, C.~Mues, J.~Vanthienen, and B.~Baesens.
\newblock An empirical evaluation of the comprehensibility of decision table,
  tree and rule based predictive models.
\newblock \emph{Decision Support Systems}, 51\penalty0 (1):\penalty0 141--154,
  2011.

\bibitem[Lundberg and Lee(2017)]{Lundberg2017}
S.~M. Lundberg and S.~Lee.
\newblock A unified approach to interpreting model predictions.
\newblock In \emph{Advances in Neural Information Processing Systems},
  volume~30, pages 4765--4774, 2017.

\bibitem[MacQueen(1967)]{Macqueen1967}
J.~MacQueen.
\newblock Some methods for classification and analysis of multivariate
  observations.
\newblock In \emph{Proceedings of the fifth Berkeley symposium on mathematical
  statistics and probability}, volume~1, pages 281--297, 1967.

\bibitem[Meteier et~al.(2019)Meteier, Capallera, Angelini, Mugellini, Khaled,
  Carrino, De~Salis, Galland, and Boll]{Meteier2019}
Q.~Meteier, M.~Capallera, L.~Angelini, E.~Mugellini, O.~A. Khaled, S.~Carrino,
  E.~De~Salis, S.~Galland, and S.~Boll.
\newblock Workshop on explainable ai in automated driving: a user-centered
  interaction approach.
\newblock In \emph{Proceedings of the 11th International Conference on
  Automotive User Interfaces and Interactive Vehicular Applications}, pages
  32--37, 2019.

\bibitem[Molnar(2020)]{Molnar2020}
C.~Molnar.
\newblock \emph{Interpretable machine learning: A guide for making black box
  models explainable}.
\newblock Leanpub, 2020.
\newblock URL \url{https://christophm.github.io/interpretable-ml-book/}.

\bibitem[Molnar et~al.(2018)Molnar, Bischl, and Casalicchio]{Molnar2018}
C.~Molnar, B.~Bischl, and G.~Casalicchio.
\newblock iml: An {R} package for interpretable machine learning.
\newblock \emph{Journal of Open Source Software}, 3\penalty0 (26):\penalty0
  786, 2018.

\bibitem[Molnar et~al.(2020)Molnar, Casalicchio, and Bischl]{Molnar2020b}
C.~Molnar, G.~Casalicchio, and B.~Bischl.
\newblock Interpretable machine learning--a brief history, state-of-the-art and
  challenges.
\newblock \emph{arXiv preprint arXiv:2010.09337}, 2020.

\bibitem[NAIC(2012)]{NAIC}
NAIC.
\newblock \emph{Model 777 - Guideline 1775 - Guideline 1780 - Product filing
  review handbook}, 2012.
\newblock URL \url{https://naic.org/prod_serv_model_laws.htm}.

\bibitem[Nelder and Wedderburn(1972)]{Nelder1972}
J.~A. Nelder and R.~W.~M. Wedderburn.
\newblock Generalized linear models.
\newblock \emph{Journal of the Royal Statistical Society: Series A (General)},
  135\penalty0 (3):\penalty0 370--384, 1972.

\bibitem[OECD(2020)]{OECD}
OECD.
\newblock \emph{The Impact of Big Data and Artificial Intelligence (AI) in the
  Insurance Sector}, 2020.
\newblock URL
  \url{https://oecd.org/finance/Impact-Big-Data-AI-in-the-Insurance-Sector.htm}.

\bibitem[O'Neil(2016)]{Oneil2016}
C.~O'Neil.
\newblock \emph{Weapons of math destruction: How big data increases inequality
  and threatens democracy}.
\newblock Crown Publishing Group, New York, 2016.

\bibitem[PRIIPs(2014)]{PRIIPs}
PRIIPs.
\newblock {R}egulation ({EU}) 1286/2014 of the {E}uropean {P}arliament and of
  the {C}ouncil of 26 {N}ovember 2014 on key information documents for packaged
  retail and insurance-based investment products.
\newblock \emph{O.J. (L 352)}, 1:\penalty0 1--23, 2014.

\bibitem[Ribeiro et~al.(2016)Ribeiro, Singh, and Guestrin]{Ribeiro2016}
M.~T. Ribeiro, S.~Singh, and C.~Guestrin.
\newblock Why should {I} trust you? {E}xplaining the predictions of any
  classifier.
\newblock In \emph{Proceedings of the 22nd ACM SIGKDD international conference
  on knowledge discovery and data mining}, pages 1135--1144. ACM, 2016.

\bibitem[Ribeiro et~al.(2018)Ribeiro, Singh, and Guestrin]{Ribeiro2018}
M.~T. Ribeiro, S.~Singh, and C.~Guestrin.
\newblock Anchors: High-precision model-agnostic explanations.
\newblock In \emph{AAAI Conference on Artificial Intelligence}, volume~18,
  pages 1527--1535. AAAI, 2018.

\bibitem[Shapley(1953)]{Shapley1953}
L.~S. Shapley.
\newblock A value for $n$-person games.
\newblock \emph{Contributions to the Theory of Games}, 2\penalty0
  (28):\penalty0 307--317, 1953.

\bibitem[Song(2019)]{Ckmeans.1d.dp}
J.~Song.
\newblock \emph{Ckmeans.1d.dp: Optimal, fast, and reproducible univariate
  clustering}, 2019.
\newblock URL \url{https://cran.r-project.org/package=Ckmeans.1d.dp}.
\newblock R package version 4.3.0.

\bibitem[{\v{S}}trumbelj and Kononenko(2010)]{Strumbelj2010}
E.~{\v{S}}trumbelj and I.~Kononenko.
\newblock An efficient explanation of individual classifications using game
  theory.
\newblock \emph{Journal of Machine Learning Research}, 11:\penalty0 1--18,
  2010.

\bibitem[{\v{S}}trumbelj and Kononenko(2014)]{Strumbelj2014}
E.~{\v{S}}trumbelj and I.~Kononenko.
\newblock Explaining prediction models and individual predictions with feature
  contributions.
\newblock \emph{Knowledge and information systems}, 41\penalty0 (3):\penalty0
  647--665, 2014.

\bibitem[Wang and Song(2011)]{Wang2011}
H.~Wang and M.~Song.
\newblock {Ckmeans.1d.dp}: optimal {K}-means clustering in one dimension by
  dynamic programming.
\newblock \emph{The R journal}, 3\penalty0 (2):\penalty0 29, 2011.

\end{thebibliography}

\end{document}